\documentclass[3p,times,authoryear]{elsarticle}
\nopreprintlinetrue

\usepackage{amssymb}
\usepackage{bm}
\usepackage{enumitem}
\usepackage{xcolor}
\usepackage{graphicx}
\usepackage{makecell}
\usepackage[outdir=./]{epstopdf}
\usepackage{caption}
\usepackage{subcaption}
\usepackage{tabularx}
\usepackage{amsfonts}
\usepackage{amsmath}
\usepackage{amssymb}
\usepackage{stfloats}
\usepackage{amsthm}
\usepackage[only,llbracket,rrbracket]{stmaryrd}
\usepackage{hyperref}
\usepackage{cleveref}
\usepackage{bbold}
\usepackage{booktabs}
\usepackage{tabularray}
\usepackage{adjustbox}
\usepackage{calc}
\usepackage{booktabs}
\usepackage{array}
\usepackage{algorithm}
\usepackage{algpseudocode}
\usepackage{multirow}
\usepackage{soul}

\usepackage{nomencl}

\setstcolor{red}

\usepackage[figuresright]{rotating}

\begin{document}

\begin{frontmatter}




\title{A reproducible comparative study of categorical kernels for Gaussian process regression, with new clustering-based nested kernels}



    \author[label1,label2]{Raphaël CARPINTERO PEREZ}
    \author[label3]{Sébastien DA VEIGA}
    \author[label2]{Josselin GARNIER}
    
    \address[label1]{Safran Tech, Digital Sciences \& Technologies, 78114 Magny-Les-Hameaux, France}
    \address[label2]{Centre de Mathématiques Appliquées, Ecole polytechnique, Institut Polytechnique de Paris, 91120 Palaiseau, France}
    \address[label3]{Univ Rennes, Ensai, CNRS, CREST - UMR 9194, F-35000 Rennes, France}


\begin{abstract}
Designing categorical kernels is a major challenge for Gaussian process regression with continuous and categorical inputs. Despite previous studies, it is difficult to identify a preferred method, either because the evaluation metrics, the optimization procedure, or the datasets change depending on the study. In particular, reproducible code is rarely available. The aim of this paper is to provide a reproducible comparative study of all existing categorical kernels on many of the test cases investigated so far. We also propose new evaluation metrics inspired by the optimization community, which provide quantitative rankings of the methods across several tasks. From our results on datasets which exhibit a group structure on the levels of categorical inputs, it appears that nested kernels methods clearly outperform all competitors.
When the group structure is unknown or when there is no prior knowledge of such a structure, we propose a new clustering-based strategy using target encodings of categorical variables. We show that on a large panel of datasets, which do not necessarily have a known group structure, this estimation strategy still outperforms other approaches while maintaining low computational cost.
\end{abstract}


\end{frontmatter}


\section{Introduction}
\label{sec:intro}

Computer experiments have become an invaluable tool in many engineering and industrial applications due to their ability to approximate expensive physical experiments or computational models \citep{santner2003design}. The study of complex systems can involve different objectives, ranging from sensitivity analysis \citep{da2021basics} to design of experiments \citep{gramacy2020surrogates}, uncertainty propagation \citep{girard2004approximate}, calibration \citep{kennedy2001bayesian} and black-box optimization \citep{jones1998efficient}. In this context, a key objective is to create a cheap-to-evaluate surrogate model from a true underlying function while providing good predictivity and characterizing sensible predictive uncertainty. To achieve this goal, one often resorts to Gaussian processes \citep{rasmussen2003gaussian} also called Kriging models \citep{krige1951statistical}, known for their ability to generalize well even when the number of observations is small. The vast majority of papers dealing with Gaussian process regression are restricted to the case where the inputs are continuous, with universally known kernels that can be used off-the-shelf. In practice, however, it is common to have access to inputs in more complicated forms, such as graphs \citep{perez2024gaussian}, meshes \citep{kabalan2025mmgp}, images \citep{yin2022uncertainty} or even probability distributions \citep{meunier2022distribution}, which all require specific kernels to be defined. In this study, we specifically address the case of mixed continuous and categorical inputs. This problem is not new, and myriad of applications can be found in fields such as chemical and antibody design \citep{gomez2018automatic, khan2023toward}, aircraft design \citep{beauthier2014hypersonic, jesus2021surrogate, bartoli2019adaptive, saves2022bayesian}, neural architecture search \citep{wistuba2019}, vegetative filter strips \cite{lauvernet2020metamodeling} and even cooking \citep{kochanski2017bayesian}. 

The term categorical is used in different senses in the literature. A first distinction must be made between categorical variables, otherwise known as nominal, and ordinal variables, which can take a finite number of ordered values. Some approaches focus specifically on binary variables, used for instance in boolean feasibility problems, while others handle categorical variables when they form strings \citep{moss2020boss, grosnit2022boils}. It should also be noted that some papers deal with overly specific cases, such as category-specific continuous variables \citep{gopakumar2018algorithmic,nguyen2020bayesian}. Finally, one of the most common applications of mixed-variable Gaussian processes is Bayesian optimization: the focus is therefore on optimization itself, sometimes neglecting the kernel's predictive performance. In such setting the kernel is often under-parameterized or simply ignores the categorical nature of inputs.

Although some reviews exist \citep{zhang2015computer,pelamatti2020mixed,roustant2020group, saves2024high} and numerous approaches have been proposed, we argue that it is difficult to make a clear recommendation on the categorical kernels to be used in practice. Indeed, papers are very rarely accompanied by codes, and reproducibility of some results is often cumbersome. In addition, optimization of kernel parameters is very sensitive to implementation details, from the choice of the optimizer, the number of multi-start trials or the bounds. Without available code, trust in previous benchmarks may be problematic. Our aim here is to close this gap by providing the practitioner a guide to choosing a categorical kernel, with reproducible experiments on several tasks. Our contribution is three-fold. First, we provide an overview of recent categorical kernels, with a critical look at some of the kernels used in Bayesian optimization. In particular, we point out that several kernels actually boil down to a naive one-hot encoding transformation. Second, all these kernels are implemented in the accompanying code, with optimization strategies clearly stated and finally compared on common datasets, with two optimization settings: one where we use the default values of optimization control parameters (as in most Gaussian process software), and one where we consider much less restrictive values (e.g. we allow for larger numbers of iterations) in order to increase the chance to reach convergence. This provides a ranking of the methods, which is used to identify the best performing methods in an objective way, while also considering their computational cost. In the case of multiple categorical variables, we also evaluate the accuracy when they are considered separately or when product variables are used. Finally, we focus on the identification of possible groups within levels when they are unknown or when no such prior assumption exists, in order to be able to use nested kernels \citep{roustant2020group} in a general context. In the latter, it is suggested that these groups may be selected by clustering the covariance matrices obtained after running another method in a preliminary step. However, numerical experiments were not carried out to explore this idea, and the coupling with another supervised approach may be a computational bottleneck. We propose instead to use a procedure that does not require any prior training. This new procedure represents each level thanks to the distribution of the output conditioned by the categorical variable taking that specific level, thus making it possible to define a similarity between the levels informed by the output variable, followed by a clustering step. We show that this approach outperforms other methods while maintaining low computational cost. To summarize, our main objectives are the following:
\begin{itemize}
    \item Present a critical review of the most popular strategies and their variants, including group kernels, hyperspheres and latent encodings
    \item Propose an extensive benchmark in Python with reproducible results, which accounts for the computational time
    \item Investigate different initialization strategies for the groups in nested kernels when groups are unknown
\end{itemize}

The article is organized as follows. We start by introducing Gaussian process regression in the special case of mixed-variable inputs and review classical categorical kernels in Section \ref{sec:overview}. We then divide our experiences into two distinct parts. Section \ref{sec:experiments1} focuses on datasets with known groups, while Section \ref{sec:experiments2} highlights datasets with no known groups and discusses strategies for identifying groups in this case. 

\section{Categorical variables in Gaussian process regression}
\label{sec:overview}

\subsection{Notations and overview}

Let us consider the case where we observe both $n\in \mathbb{N}$ continuous variables with values in $\mathbb{R}$, and $m$ categorical variables with respectively $C_1, ..., C_m$ levels (also called categories or modalities). Without loss of generality, we denote the sets of levels $\mathcal{Z}_i = \{1, \cdots , C_i\} = \llbracket C_i \rrbracket$ for variable $i$, so that the space of categorical variables is written as $\mathcal{Z} = \prod\limits_{i=1}^m \mathcal{Z}_i$. The input space is thus $\mathcal{W} = \mathbb{R}^{n}\times \mathcal{Z}$. We consider the task of learning a function $f : \mathcal{W} \rightarrow \mathbb{R}$ from a dataset $\mathcal{D}$ of $N$ observations $\mathcal{D} = \{ (W^{(i)}, y^{(i)}) \}^{N}_{i=1}$ where $W^{(i)} = (x^{(i)}, z^{(i)})$, with noisy observations $y^{(i)} = f(W^{(i)}) + \epsilon^{(i)}$ for $i \in \llbracket N\rrbracket$ at input locations $\mathbf{W} = (W^{(i)})_{i=1}^N$, where $\epsilon^{(i)} \sim \mathcal{N}(0,\eta^2)$ is an i.i.d. Gaussian additive noise. We denote $\mathbf{y} = (y^{(i)})_{i=1}^N$ and $\mathbf{f_*}:=(f({W}_*^{(i)}))_{i=1}^{N^*}$ the values of $f$ at new test locations $\mathbf{W}_*=({W}_*^{(i)})_{i=1}^{N^*}$. A constant-mean Gaussian process (GP) prior is placed on the function $f$. To simplify the notations, we will use a zero-mean here. It follows that the joint distribution of the noisy function values $\mathbf{y}$ at the observed locations and the function values $\mathbf{f}_*$ at the test locations writes (see, \textit{e.g.}, \citet{rasmussen2003gaussian})
\begin{equation}
    \begin{bmatrix}
    \mathbf{y}\\
    \mathbf{f}_*
    \end{bmatrix}
    \sim \mathcal{N} 
    \left( 
    \mathbf{0}, 
    \begin{bmatrix}
    \mathbf{K} + \eta^2 \mathbf{I} & \mathbf{K}_*^\top\\
    \mathbf{K}_* & \mathbf{K}_{**}
    \end{bmatrix}
    \right)\,,
\end{equation}

where $\mathbf{K}$, $\mathbf{K}_{**}$, $\mathbf{K}_{*}$ are the train, test and test/train Gram matrices, respectively.

The posterior distribution of $\mathbf{f}_*$, obtained by conditioning the joint distribution on the observed noisy data, is also Gaussian: $\mathbf{f}_* | \mathbf{W}, \mathbf{y}, \mathbf{W}_* \sim \mathcal{N}(\mathbf{\bar{m}},\mathbf{\bar{\Sigma}})$ with mean and covariance given by
\begin{align}
\mathbf{\bar{m}}&=\mathbf{K}_*(\mathbf{K}+\eta^2\mathbf{I})^{-1} \mathbf{y}\,, \\
\mathbf{\bar{\Sigma}}&=\mathbf{K}_{**}-\mathbf{K}_* (\mathbf{K}+\eta^2 \mathbf{I})^{-1}\mathbf{K}_*^\top\,.
\end{align}
The mean of the posterior distribution is used as a predictor, and predictive uncertainties can be obtained through the posterior covariance matrix. 

Since prediction formulas are in closed-form, the cornerstone of GP regression lies in the choice of the symmetric kernel function $k: \mathcal{W} \times \mathcal{W} \rightarrow \mathbb{R}$, which must be positive definite (also called valid kernel). 

Although some approaches try to build a kernel on $\mathcal{W}$ directly \citep{oh2021mixed}, the standard way to tackle the problem is based on the construction and combination of separate kernels on the continuous and categorical parts. Suppose we have access to a valid kernel for the continuous variables $k_{\text{cont}}: \mathbb{R}^n \times \mathbb{R}^n$, and one for the categorical variables $k_{\text{cat}}: \mathcal{Z}\times \mathcal{Z}$. Usual valid combinations include the product, sum, or ANOVA. 

\begin{figure}
\begin{center}
\includegraphics[width=0.8\linewidth]{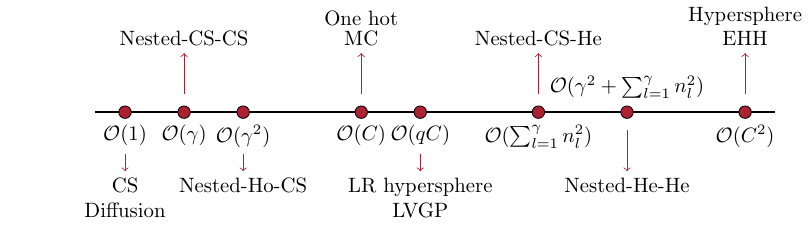}
\end{center}
\caption{Number of parameters of several categorical kernels. $C$ : number of levels, $q$: latent dimension (for encoding-based or low-rank approaches), $\gamma$: number of groups (for nested kernels), $n_l$: number of levels per group (for nested kernels). Remark that $C=\sum_{l=1}^\gamma n_l$  so that we always have $C \leq  \sum_{l=1}^\gamma n_l^2$ and $\sum_{l=1}^\gamma n_l^2 \leq C^2 \leq \gamma \sum_{l=1}^\gamma n_l^2$.
\label{fig:chronology}}
\end{figure}

RBF and Mat\'ern kernels are reference choices when variables are continuous. For categorical variables, it is customary to consider each categorical variable separately, and to combine them after. Although additive kernels proposed by \cite{deng2017additive} offer an interesting alternative with the sum, the results of \cite{roustant2020group} show that they have inferior performance, so we limit our study to the most common product kernels. We thus seek to define a categorical kernel $k_{\text{cat}}(z, z') = \prod\limits_{i=1}^m k_{\text{cat},i}(z_i, z'_i)$ where $z=(z_1, ..., z_m) \in \mathcal{Z}$, $z'=(z'_1, \cdots, z'_m) \in \mathcal{Z}$. In the next section, in order to simplify notations, we consider that $m=1$, $\mathcal{Z}=\llbracket C \rrbracket$ and $k_{\text{cat}}: \mathcal{Z}\times \mathcal{Z} \rightarrow \mathbb{R}$. We identify two ways of constructing such kernels. The first one is to create a data representation, or encoding, which is then plugged into a traditional continuous kernel. The second, as described by \cite{zhou2011simple}, relies on the fact that a categorical variable is characterized by a finite number of levels and thus can be described by a finite number of covariance values that we denote by $\mathbf{T} = (T_{z,z'}) \in \mathbb{R}^{C\times C}$, such that $k_{\text{cat}}(z,z') = T_{z,z'}$. By choosing a suitable parameterization of the covariance matrix, the positive definiteness of $k_{\text{cat}}$ follows from the positive definiteness of $\mathbf{T}$. Each method has its own number of parameters to be inferred from the data, see Figure \ref{fig:chronology} for a summary of the main approaches described in the following subsections.

\subsection{Encoding}

\paragraph{One-hot encoding}~\\
A categorical variable with $C$ levels can be represented by a vector of size $C$ composed of zeros and ones. $z \in \llbracket C \rrbracket$ can be written as $E(z) \in \{0,1\}^{C}$ where $E(z)_i = \mathbb{1}_{\{i=z\}}$ for $i=1, \cdots, C$. The encoded variable is then plugged into continuous kernels, such as the RBF:
$k_{\text{cat}}(z,z') = \prod\limits_{i=1}^C e^{-\frac{((E(z))_i-(E(z'))_i)^2}{2\theta_i^2}}$ where $\theta=(\theta_1, \cdots, \theta_C)\in (0, \mathbb{+\infty})^C$. Relaxation approaches are often used in Bayesian optimization \citep{garrido2020dealing, karlsson2020continuous,halstrup2016black, kim2022combinatorial}, where one-hot encoded variables are further allowed to take values in $[0,1]^C$. However, note that these kernels have the trivial expression: 
\begin{equation}
k_{\text{cat}}(z,z') = \mathbb{1}_{\{z=z'\}} + e^{-\frac{1}{2}(\theta^{-2}_z+\theta^{-2}_{z'}) } \mathbb{1}_{\{z\neq z'\}}, \label{eq:one-hot-kernel}
\end{equation}
which is the sum of a diagonal matrix and a rank-one matrix. This shows the limited representational power of one-hot encoding.

\paragraph{Latent variables}~\\
Instead of using a fixed representation, Latent Variable Gaussian Process (LVGP) \citep{zhang2020latent} proposes to infer a representation in dimension $q$ where $1\leq q<C$. Indeed, it is common in engineering that categorical variables are explained by unobserved continuous ones (e.g., a material type is characterized by several physical properties). Such a representation is defined by a mapping $\phi: \llbracket C\rrbracket \rightarrow \mathbb{R}^q$, parameterized by $qC - \frac{q(q+1)}{2}$ variables:
\begin{equation}
\begin{aligned}
 \phi(1) &= (0, \phantom{\theta_{i,1}} 0, \phantom{\theta_{i,1}} \cdots \phantom{0 \theta_{i,i-1}, 0 0, 0 \cdots, 0 0} 0)\\
 \phi(2) &= (\theta_{2,1}, \phantom{0} 0, \phantom{\theta_{i,1}} \cdots \phantom{0 \theta_{i,i-1}, 0 0, 0 \cdots, 0 0} 0) \\
 \phi(i) &= (\theta_{i,1}, \phantom{0} \theta_{i,2}, \phantom{0} \cdots, \phantom{0} \theta_{i,i-1}, \phantom{0} 0, \phantom{0} \cdots \phantom{00} 0) &\mbox{ for } 2\leq i \leq q\\
 \phi(i) &= (\theta_{i,1}, \phantom{0} \theta_{i,2}, \phantom{0} \cdots\phantom{0 \theta_{i,i-1}, 0 0, 0 \cdots, 0 0} \theta_{i,q}) &\mbox{ for } i>q.\\
\end{aligned}
\end{equation}
Finally, LVGP relies on a continuous RBF kernel between these low-dimensional embeddings as follows:
$$
k_{\text{cat}}(z,z') = \exp(- ||\phi(z)-\phi(z')||^2_2).
$$
According to \citet{zhang2020latent}, while choosing $q=1$ lacks expressivity and $q=C-1$ is too large, using $q=2$ is sufficient in many practical cases. The LVGP kernel was successfully used in \cite{cuesta2022comparison}. In the case where several categorical variables are observed, we can also mention the work of \citep{oune2021latent}, where they propose a common latent representation for all possible levels in the product space instead of defining a latent representation for each categorical variable separately. Note also that another way to learn latent representations is to use deep generative models \citep{notin2021improving, deshwal2021combining, maus2022local}. 

\subsection{Covariance matrix parameterization}
In the covariance matrix parameterization framework, recall that $k^{\text{cat}}(z,z') = T_{z,z'}$ where $\mathbf{T}$ is the $C\times C$ kernel matrix, with unknown entries to be inferred. Symmetry and positive definiteness properties of $\mathbf{T}$ are usually exploited to find simple parameterizations \citep{pinheiro1996unconstrained}, but the most popular ones involve spectral or Cholesky decompositions, and sparse representations which we detail below.

\paragraph{Spectral decomposition}~\\
Since $\mathbf{T}$ needs to be symmetric, one can use the spectral decomposition $\mathbf{T} = \mathbf{P}\mathbf{D}\mathbf{P}^\top$, where $\mathbf{P}$ is orthogonal and $\mathbf{D}$ is diagonal. Then, the problem boils down to finding a suitable parameterization of the orthogonal matrix $\mathbf{P}$. These include, for example, the Eulerian angles or the Cayley and the Householder transforms \citep{khuri1989parameterization, shepard2015representation}. These representations require learning $C(C-1)/2$ parameters, plus the $C$ parameters for the diagonal matrix parameters. These representations are however less popular than the following ones, since they only exploit symmetry.

\paragraph{Cholesky decomposition}~\\
Rather than the spectral decomposition, it is more common to use the Cholesky decomposition, which applies to symmetric and positive definite matrices \citep{rapisarda2007parameterizing}. The Cholesky decomposition writes as $\mathbf{T} = \mathbf{L}\mathbf{L}^\top$ where $\mathbf{L}$ is a lower triangular matrix. As described in \citet{zhou2011simple}, such lower triangular matrices can be parameterized by the following \textbf{heteroscedastic hypersphere} (He) decomposition, also called unrestrictive covariance:
\begin{equation}
\label{eq:hypersphere2} 
\begin{aligned}
L_{1,1} &= \theta_{1,0} \\
L_{i,1} &= \theta_{i,0} \cos(\theta_{i,1})  \mbox{ for } 2\leq i \leq C\\
L_{i,i} &= \theta_{i,0} \prod_{l=1}^{i-1} \sin(\theta_{i,l}) \mbox{ for } 2\leq i \leq C\\
L_{i,j} &= \theta_{i,0} \cos(\theta_{i,j}) \prod_{l=1}^{j-1} \sin(\theta_{i,l}) \mbox{ for } 2\leq j<i \leq C\\
\end{aligned}
\end{equation}
where $\theta_{i,j} \in (0,\pi)$ for $1\leq j < i \leq C$ and $\theta_{i,0} >0$ for $1\leq i \leq C$. The intuition behind this parameterization is that each level can be represented as a point on the surface of the $C$-dimensional hypersphere. By default, it depends on $\frac{C (C+1)}{2}$ parameters. The number of parameters can be further reduced to $\frac{C (C-1)}{2}$ when we assume that all levels share the same variance, i.e. $\theta_{i,0} = 1$ for all $i$: this is the \textbf{homoscedastic hypersphere} (Ho).\\
Furthermore, in this setting the entries of $\mathbf{T}$ take values in $(-1,1)$, thus allowing negative correlations. Restricting $\theta_{i,j} \in (0,\frac{\pi}{2})$ for $1\leq j < i \leq C$, all entries of $\mathbf{T}$ take value in $(0,1)$, thus exhibiting only positive correlations.

The hypersphere parameterization (He or Ho, with or without negative correlations) has the advantage of being very rich, but has a very large number of parameters to optimize, which can be prohibitive when the number of levels increases, in particular when using values for the optimizer control parameters that are adapted to such dimensionality. This explains why sparse variants have recently emerged.

\paragraph{Low-rank matrices}~\\
Similarly to the LVGP approach, it is possible to assume that the data live in a lower rank space such that $\textbf{T} = \mathbf{U}\mathbf{U}^\top$,
where $\mathbf{U}$ is a $C \times  q$ matrix, and $q<C$. Starting from the hypersphere parameterization, and fixing some angles to $0$, \cite{kirchhoff2020gaussian} build a representation with $(q-1){(C-\frac{q}{2})}$ parameters.

\paragraph{Multiplicative covariance}~\\
Another kernel that is present in the literature is the multiplicative covariance kernel \citep{mcmillan1999analysis, qian2008gaussian}. Its form is as follows:
$$
T_{z,z'} = \mathbb{1}_{\{z=z'\}} + \exp(-\theta_{z} - \theta_{z'}) \mathbb{1}_{\{z\neq z'\}}
$$
where $\theta_i \in [0,+\infty)$ for $1\leq i \leq  C$. Note the striking similarly with Equation (\ref{eq:one-hot-kernel}): this multiplicative kernel in fact amounts to choosing a one-hot encoding of the variables, and then applying a continuous exponential kernel. As expected, such under-parameterization fails to capture commonly occurring correlation structures when $C\geq 4$ as pointed out by \cite{zhang2015computer} and \cite{zhang2020latent}.

\paragraph{Parametric exponentiation}~\\
\cite{saves2023mixed} generalize the multiplicative covariance by introducing a symmetric positive definite matrix $\boldsymbol{\tau} \in \mathbb{R}^{C\times C}$ characterizing the correlations between all the levels of the categorical variable:
\begin{equation}
\label{eq:saves} 
\begin{aligned}
\mathbf{T}_{z,z'}= \exp\left( - \tau_{z,z} - \tau_{z',z'} - 2 \tau_{z,z'} \right) \mathbb{1}_{\{z\neq z'\}} + \mathbb{1}_{\{z = z'\}}.
\end{aligned}
\end{equation}
$\boldsymbol{\tau}$ is further parameterized using the hypersphere decomposition (\ref{eq:hypersphere2}), with parameters $\theta_{i,j}\in (0,\frac{\pi}{2})$ for $1\leq j<i \leq C$, $\theta_{i,0}=1$ for $1\leq i \leq C$ and a small tolerance $0<\epsilon \ll 1$. This amounts to defining $\tau_{i,i} = \theta_{i,i} \geq 0$ for $1\leq i \leq C$ and $\tau_{i,j} = \frac{\log(\epsilon)}{2}\left( (\mathbf{L}\mathbf{L}^T)_{i,j} -1 \right) $ for $i\neq j$. With this parameterization, called Fully Exponential (FE), the covariance matrix $\mathbf{T}$ is positive definite, with values in $[\epsilon, 1)$ and uses $\frac{C (C+1)}{2}$ parameters. \cite{saves2023mixed} also propose an alternative way to build the correlation matrix (\ref{eq:saves}) by imposing that $\theta_{i,i} = 0$ for $1\leq i\leq C$, thus leading to:
\begin{equation*}
    T_{z,z'} = \exp(-2\tau_{z,z'})\mathbb{1}_{\{z\neq z'\}} + \mathbb{1}_{\{z = z'\}}.
\end{equation*}
They call it \textbf{exponential homoscedastic hypersphere} (EHH), and it depends on $\frac{C (C-1)}{2}$ parameters.

\paragraph{Compound symmetry}~\\
Unlike hypersphere parameterizations, which have a number of parameters growing quadratically with the number of levels, some models propose to describe the covariance matrices with a number of parameters independent of the number of levels. For example, the simplest models consist in setting a constant covariance value between levels. The \textbf{compound symmetry} (CS) model \citep{katz2011multivariable}, also called exchangeable covariance, writes $\mathbf{T}$ as a function of two values, $c$ for covariance and $v$ for variance between levels:
$$T_{z,z'} = v \mathbb{1}_{\{z = z'\}} + c \mathbb{1}_{\{z \neq z'\}}.$$
This defines a valid kernel as long as $v> 0$ and $\frac{c}{v} \in (-\frac{1}{C-1}, 1)$. 
Subcases are occasionally found, such as the Aitchen-Aitken kernels \citep{aitchison1976multivariate, watanabe2023}, fixing $v=1-b$ and $c=\frac{b}{C-1}$. More surprisingly, we can also mention the kernels of \cite{oh2019} for $m$ categorical variables. Their approach proposes to represent the combinatorial space thanks to a graph Cartesian product of combinatorial graphs, and then to use a diffusion kernel \citep{kondor2002diffusion} on the factor graph. However, without any further knowledge about the graph structure of the combinatorial space, the diffusion kernel for complete graphs subsequently has a closed form expression which does not require any spectral decomposition:
$$  T_{z,z'} = \left(\frac{1+(C-1)e^{-\beta C}}{C}\right) \mathbb{1}_{\{z=z'\}}  + \left(\frac{1-e^{-\beta C}}{C}\right) \mathbb{1}_{\{z\neq z'\}}$$
where $\beta>0$. Hybrid kernels of \cite{deshwal2021bayesian} also take a very similar form, while Mercer features of \citep{deshwal2021mercer} also use these diffusion kernels and derive an explicit representation of $m$ binary variables. After rewriting, it is clear that these kernels are equivalent to the simple compound symmetry kernel, although this is never mentioned explicitly in the literature.

\paragraph{Nested kernels}~\\
The \textbf{nested kernels}, also called group kernels or generalized compound symmetry (GCS), were proposed by \cite{roustant2020group} as a generalization of the group compound symmetry model of \citep{qian2008gaussian}. In particular, they studied validity conditions of such kernels, which previously had no theoretical guarantees. The main assumption for nested kernels is that the levels can be divided into different groups. They consist in building a sparse representation of $\textbf{T}$ which involves two types of blocks: within-group covariances and covariances between groups. Suppose $\llbracket C \rrbracket$ is partitioned into $\gamma$ groups $\mathcal{G}_1, \cdots, \mathcal{G}_\gamma$ of respective sizes $n_l, 1\leq l \leq\gamma$. The nested kernel writes as follows:
$$
\mathbf{T} = 
\begin{pmatrix}
\mathbf{W}_1 & \mathbf{B}_{1,2} & \cdots & \mathbf{B}_{1,\gamma} \\
\mathbf{B}_{2,1} & \mathbf{W}_2 & \ddots & \mathbf{B}_{2,\gamma} \\
\vdots & \ddots & \ddots & \vdots \\
\mathbf{B}_{\gamma,1} & \dots & \mathbf{B}_{\gamma,\gamma-1} & \mathbf{W}_{\gamma} \\
\end{pmatrix}
$$
where the diagonal blocks $\mathbf{W}_l$ are $n_l\times n_l$ matrices containing the within-group covariances and $\mathbf{B}_{l,l'}$ are constant matrices containing between-group covariances. The authors put forward simple conditions to verify the positive definiteness of $\mathbf{T}$: $\mathbf{W}_l$ and $\mathbf{W}_l$ minus its mean need to be positive semidefinite for all $l$ and the $\gamma\times \gamma$ matrix $\tilde{\mathbf{T}}$ obtained by averaging each block of $\mathbf{T}$ needs to be positive definite. In order to meet these conditions, they propose a way to parameterize valid GCS block matrices by a family of covariance matrices of smaller size. By using $\mathbf{B}^*$ a $\gamma \times \gamma$ matrix of between-group means covariance, and $\mathbf{M}_l$ a centered covariance matrix of size $(n_l-1)\times(n_l-1)$ for $l=1,...,\gamma$, they form the following block matrix:\\
\begin{align*}
    \mathbf{W}_l &= \mathbf{B}^*_{l,l} \mathbf{J}_{n_l, n_l} + \mathbf{A}_l \mathbf{M}_l \mathbf{A}_l^T\\
    \mathbf{B}_{l,l'} &= \mathbf{B}^{*}_{l,l'} \mathbf{J}_{n_l, n_{l'}}
\end{align*}
where $\mathbf{J}_{a,b}$ is the $a \times b$ matrix of ones and $\mathbf{A}_l$ is a $n_l\times (n_l-1)$ matrix whose columns form an orthonormal basis of $1_{n_l}^{\bot}$.
Different levels of sparsity can be obtained with different choices of generator matrices as detailed in Table \ref{tab:gcs_param} taken from \cite{roustant2020group}. 

\begin{table*}[h!]
\setlength{\tabcolsep}{8pt}
\caption{Parameterization details for some valid GCS block covariance matrices (the two first columns give the parametric setting).}
\label{tab:gcs_param}
\centering
\begin{tabular}{|ccccc|} 
 \toprule
$\mathbf{M}_l$ & $\mathbf{B}^*$ & $\mathbf{W}_l$ & $\mathbf{B}_{l,l'}$ & Number of parameters\\
 \midrule
 $v_{\lambda_l} \mathbf{I}_{n_l-1}$ & $\textrm{CS}(v,c)$ & $\textrm{CS}(v_l, c_l)$ & $c$ & $\gamma+2$\\ 
 $v_{\lambda_l} \mathbf{I}_{n_l-1}$ & Ho/He & $\textrm{CS}(v_l, c_l)$ & $c_{l,l'}$ & $\frac{\gamma(\gamma+3)}{2}$\\ 
 Ho/He & $\textrm{CS}(v,c)$ & Ho/He & $c$ & $\sum\limits_{l=1}^\gamma \frac{n_l (n_l+1)}{2} + 2$\\
 Ho/He & Ho/He & Ho/He & $c_{l,l'}$ & $ \sum\limits_{l=1}^\gamma \frac{n_l (n_l+1)}{2} + \frac{\gamma(\gamma+1)}{2}$\\ 
 \bottomrule
\end{tabular}
\end{table*}

\subsection{Kernels from the Bayesian optimization litterature}
In the Bayesian optimization community, it is much less common to use kernels such as hyperspheres, while similarity functions (not always positive definite) are often used.

Hamming Embeddings via Dictionaries (HED), see \cite{deshwal2023bayesian}, select a number of discrete structures from the categorical space (the dictionary) and use them to define an ordinal embedding for high-dimensional combinatorial structures. Given a dictionary with $q\geq 1$ elements, the Hamming embedding vector of a level corresponds to the vector of Hamming distances between the level and all the elements in the dictionary. However, this approach requires access to a sufficient number of categorical variables with low number of levels to be relevant. Alternatively, for categorical inputs expressed as string data, adapted kernels are also often used \citep{grosnit2022boils, moss2020boss}.

In many papers in the Bayesian optimization framework, the Gower distance \citep{halstrup2016black, gower1971general} is very popular. For $m$ categorical variables, the Gower distance between two inputs simply counts the average number of times the categorical variables are the same. The associated Gower distance kernel plugs the Gower distance into a power exponential kernel, but the kernel is not positive definite. Similar indefinite kernels can be found in \cite{ru2020bayesian} and \cite{wan2021think}.

\subsection{Available categorical kernels in Python}
Finally, from a practitioner perspective, we discuss which categorical kernels mentioned above are easily available in Python. Most Bayesian optimization and GP toolboxes are limited to continuous inputs, like Spearmint \citep{snoek2012}, GPyTorch \citep{gardner2018gpytorch}, GPy \citep{gpy2014}, BoTorch \citep{balandat2020botorch}. The OpenTurns software only features LVGP \citep{baudin2015open}. The MCBO framework of \citep{dreczkowski2024framework} only considers under-parameterized kernels based on the Gower distance or the diffusion kernels, HED or kernels for string data. The SMT2 toolbox \citep{saves2024smt} is the most complete as it offers continuous relaxation, Gower distance, homoscedastic hypersphere and exponentiation homoscedastic hypersphere. As a sidenote, the R package kergp \citep{deville2024package} considers hypersphere kernels and nested kernels. But all in one, it is obvious that the existing implementations only offer a very limited set of choices.

\section{Implementation and experiments with known group structure}
\label{sec:experiments1}

In this section, we describe the kernels considered in our experiments, and detail our implementation choices. We then describe our experimental protocol, and first evaluate methods on test cases with a known group structure. Experiments without such a structure will be discussed in Section \ref{sec:experiments2}. We start by recalling the list of kernels used in all experiments in Table \ref{tab:kernels}, and those used only in the known-group setting in Table \ref{tab:kernels1}. The code needed to reproduce the experiments can be found at the following URL: \url{https://gitlab.com/drti/cat_gp/}.

\begin{table}[ht!]
\caption{List of methods/kernels for all experiments.}
\small
\centering
\begin{tabular}{|llc|}
\toprule
\textbf{Category} & \textbf{Name} & \textbf{Description} \\
\midrule
\multirow{7}{*}{Hypersphere}
  & Ho & Homoscedastic, only positive correlations  \\
  & Ho\_NC & Homoscedastic, allowing negative correlations  \\
  & He & Heteroscedastic, only positive correlations  \\
  & He\_NC & Heteroscedastic, allowing negative correlations  \\
  & Ho\_2 & Homoscedastic rank 2, only positive correlations  \\
  & Ho\_3 & Homoscedastic rank 3, only positive correlations  \\
  & EHH & Exponential Homoscedastic Hypersphere  \\
\midrule
LVGP & LVGP & Latent variables (dimension 2)  \\
CS & CS & Compound symmetry  \\
One-hot & one\_hot & One-hot encoding  \\
No cat & no cat & Only continuous variables  \\
\bottomrule
\end{tabular}
\label{tab:kernels}
\end{table}

\begin{table}[ht!]
\caption{List of kernels for experiments with known group structure only.}
\small
\centering
\begin{tabular}{|llc|}
\toprule
\textbf{Category} & \textbf{Name} & \textbf{Description} \\
\midrule
\multirow{5}{*}{\shortstack{Nested Kernels \\ Known groups}} 
  & & \textbf{Between} / \textbf{Within} \\
  & Nested\_CS\_He & CS / Heteroscedastic \\
  & Nested\_He\_He & Heteroscedastic / Heteroscedastic \\
  & Nested\_Ho\_CS & Homoscedastic / CS\\
  & Nested\_CS\_CS & CS / CS\\
\bottomrule
\end{tabular}
\label{tab:kernels1}
\end{table}

\subsection{Implementation details}
\label{sec:implementation_details}
The experimental setting we describe here for Section \ref{sec:experiments1} is the same as for Section \ref{sec:experiments2}. Important details about the experiments, in particular the parameterization and bounds of all parameters to be estimated, are presented in \ref{appendix:exp_setup} to ensure reproducibility. We summarize here the most important facts:
\begin{itemize}
    \item For continuous inputs, we use an anisotropic RBF kernel with one lengthscale per dimension
    \item For categorical inputs, each considered kernel has its own set of parameters 
    \item The set of all parameters involved in the kernels (continuous and categorical) are estimated by maximizing the marginal log likelihood, except for the kernel variance which is obtained in closed-form.
    \item  Optimization is performed with the L-BFGS-B algorithm \citep{liu1989limited}, with scipy's optimize function \citep{2020SciPy-NMeth}. Crucially, and unlike previous published benchmarks, we consider two different optimization scenarios:
    \begin{enumerate}
        \item A "short" optimization setting, where we use all default options for scipy's optimize function, with a maximum of function evaluations set to 15000 and a tolerance equal to $1e^{-9}$. When the number of parameters is large, as in hypersphere kernels, such function evaluation restriction severely impacts the optimizer capacity to reach a local optimum. Indeed, gradients are estimated by finite-differences, thus requiring a potentially large number of function calls at each iteration, and subsequently limiting the allowed number of iterations.
        \item A "long" optimization setting, where we instead fix a maximum number of iterations equal to 3000, and adapt the maximum number of function calls accordingly, i.e. we set it to 3000 $\times$ (number of hyperparameters + 1 + number of line search steps). We fix the number of line search steps to 20, and lower the tolerance to $1e^{-10}$.
    \end{enumerate}
\end{itemize}

All the results given in Sections \ref{sec:experiments1} and \ref{sec:experiments2} correspond to the "long" optimization setting, but in some paragraphs we will also present experiments with the "short" optimization setting, since it will help shed light on interesting facts from a practical perspective.

\subsection{Presentation of the datasets}

Table \ref{tab:comparison1} presents the datasets used in the study with known groups. For all datasets, designs of experiments of different sizes are generated, and 50 independent replications are created to form the training sets for a given dataset. Each dataset is presented in detail in \ref{appendix:presentation_datasets}.

\begin{table*}[h]
\setlength{\tabcolsep}{6pt}
\caption{Description of datasets with known groups: number of continuous and categorical variables (numbers of levels in parenthesis), and size of training and test datasets. Groups=True means that the groups are known.}
\label{tab:comparison1}
\centering
\begin{tabular}{|ccccccc|} 
 \toprule
Name & Cont & Cat & Train & Test & Groups & Source\\
 \midrule
 $f_1$ & 1 & 1 (13) & 39/78/117/156/195 & 1001 & True & \cite{roustant2020group}\\
 $f_2$ & 1 & 1 (10) & 30/60/90/120/150 & 1000 & True & \cite{roustant2020group}\\
 Beam bending & 2 & 1 (12) & 36/72/108/144/180 & 1000 & True & \cite{roustant2020group}\\
 \bottomrule
\end{tabular}
\end{table*}

\subsection{Comparison of the methods}

The main criterion for assessing the quality of predictions is the Relative Root Mean Squared Error (RRMSE), defined for $N^*$ test ground truth scalar values $(y^{(i)})_{i=1}^{N^*}$ and their associated predictions $(\hat{y}^{(i)})_{i=1}^{N^*}$ as:
\begin{equation*}
    \text{RRMSE}\left( (y^{(i)})_{i=1}^{N^*}, (\hat{y}^{(i)})_{i=1}^{N^*} \right) = \sqrt{\frac{\sum_{i=1}^{N^*} (y^{(i)} - \hat{y}^{(i})^2}{\sum_{i=1}^{N^*} (y^{(i)} - \bar{y})^2}},
\end{equation*}
where $\bar{y} = \frac{1}{N^*} \sum_{i=1}^{N^*} y^{(i)}$ is the mean of the true values.

In the following, we report two complementary ways to evaluate the methods: individual performances on each dataset, as done usually in such benchmarks, but also performance profiles inspired by the optimization community.

\begin{figure}[h!]
\begin{center}
\includegraphics[width=0.49\linewidth]{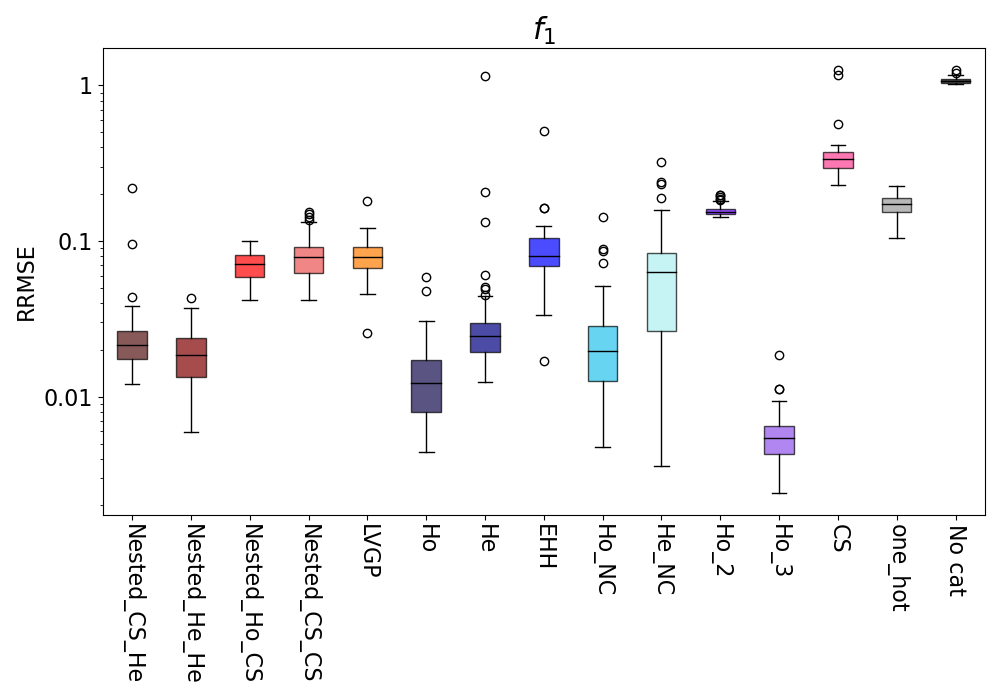}
\includegraphics[width=0.49\linewidth]{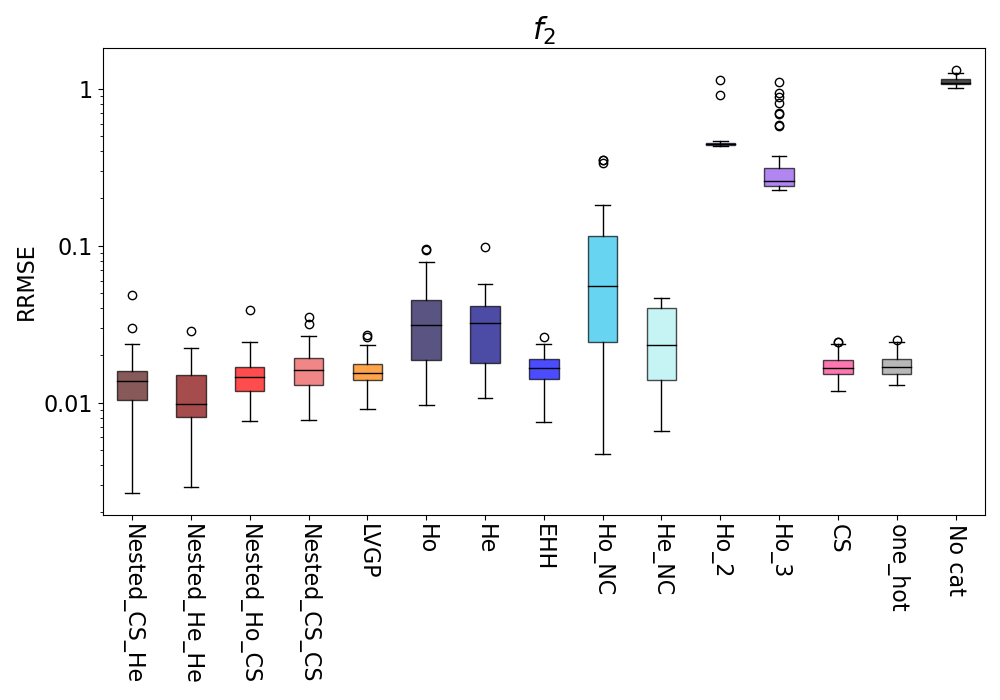}
\includegraphics[width=0.49\linewidth]{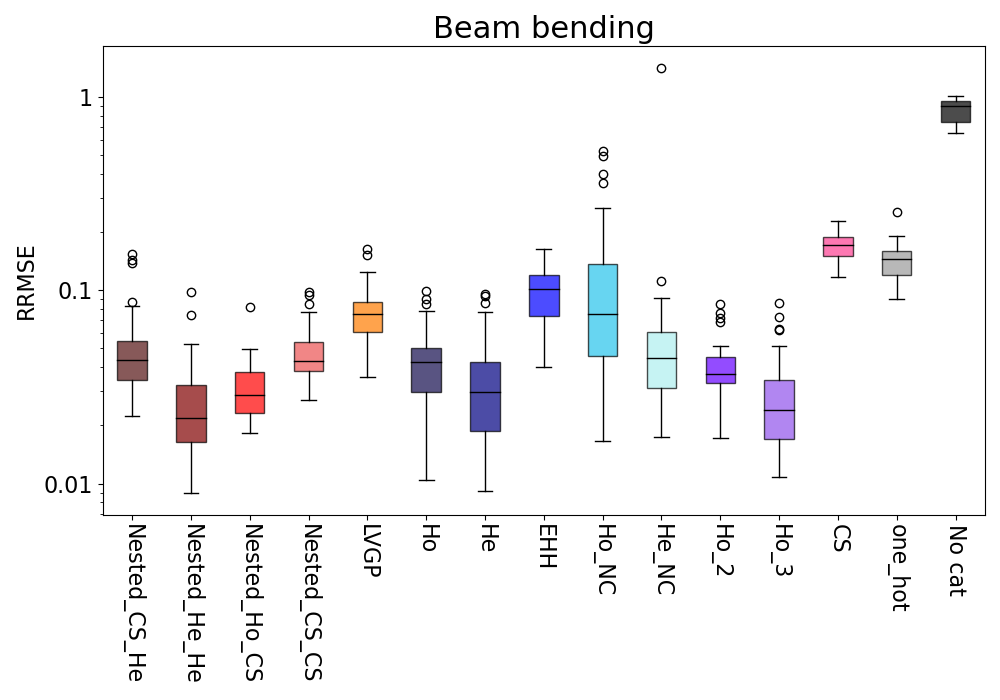}
\end{center}
\caption{Three datasets with known group structure. Boxplots of the RRSME over the 50 experiments for all methods, "long" optimization setting. \label{fig:boxplots_groups}}
\end{figure}

\paragraph{Results on each dataset}~\\ \label{sec:results_datasets1}
Since there are many dataset variants, we limit the discussion to medium-sized datasets with 6 samples per level (training size equal here to 6 times the number of levels, as there is only one categorical variable in all datasets of Table \ref{tab:comparison1}). The results for other dataset sizes are available in the repository containing the code. Figure \ref{fig:boxplots_groups} highlights the results of all compared methods on the $f_1$, $f_2$ and beam bending problems, where the groups are known.

On average, compound symmetry and one-hot encoding have low performance, except for $f_2$. Hypersphere kernels have surprisingly good performance, which is due to the long optimization setting. In some instances, low-rank homoscedastic parameterizations perform well, but there is variability depending on the test cases. Finally, consistent good performance is achieved by LVGP and EHH, but Nested\_He\_He systematically leads to better RRMSE scores.\\

In order to better analyze all the results across several replications, training set sizes and test cases, we introduce below a new way to compare methods.

\paragraph{Performance profiles}~\\
Performance profiles are inspired by profiles used in optimization benchmarks \citep{dolan2002benchmarking}, see \ref{appendix:performance_profiles} for a detailed introduction. Roughly speaking, the performance profile for a method can be seen as the cumulative distribution function for a performance metric which is the RRMSE in this section. It is a powerful visualization tool to aggregate results obtained on several datasets, with different replications and an ensemble of methods. A performance function $p_{i} : [0,1] \rightarrow [0,1]$ is associated to each method $i$. Imagine we have access to the scores of every method on all experiments on every dataset. We want to be able to determine if a method is often among the best methods on a given dataset, or if, on the contrary, it is among the worst. We should additionally take into account the scores of all experiments on each dataset. To do so, for each dataset, we first sort the scores of all methods on all experiments. Then, given a performance level $\tau \in [0,1]$, and a method $i$, $p_i(\tau)$ counts the proportion of times when an (experiment, method) pair is in the top $\tau\%$ performing methods of all experiments carried out on the given dataset. Such curves are non-decreasing, with $p_i(0)=0$ and $p_i(1)=1$. If a method outperforms the other ones on all datasets, then $p_i(\tau)$ is close to 1 even for small values of $\tau$, so the curve grows rapidly and reaches the upper left corner. Conversely, a method that underperforms on all datasets grows slowly at the beginning, approaching the lower right corner. The overall performance of method $i$ can finally be measured by the Area Under The Curve (AUC) of its performance profile plot: $\mathrm{AUC}(p_i) = \int_{0}^1 p_i(\tau)\mathrm{d}\tau$. The performance profile plots gather the curves $p_{i}$  between $0$ and $1$ for all methods, and should be read in conjunction with the individual performance figures for each dataset from Section \ref{sec:results_datasets1}.

\begin{figure}[h!]
\begin{center}
\includegraphics[width=0.9\linewidth]{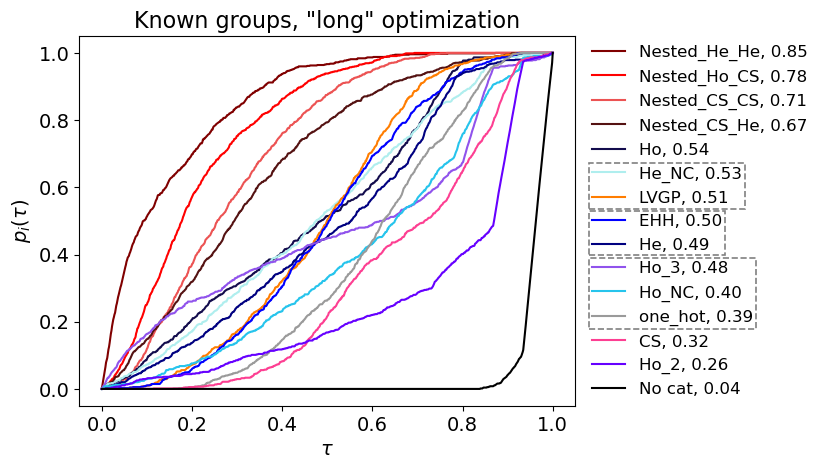}
\end{center}
\caption{Performance profiles using the RRMSE for datasets with a group structure, ”long” optimization setting. The score after the name of the method $i$ is $\mathrm{AUC}(p_i)$, methods are sorted in AUC decreasing order.}
\label{fig:profile_all_RRMSE_GD}
\end{figure}

Figure \ref{fig:profile_all_RRMSE_GD} gathers such performance profiles, with method ranked according to their AUC. In the legend, some of them are grouped, which means that their performance are not statistically different from a one-sided Wilcoxon signed rank test. We first observe that group kernels clearly outperform other kernels when there is a known group structure. The 4 types of within/between covariance structures are among the methods with the best overall performance on these datasets. Choosing hyperspheres for the between and within correlations gives better overall results compared to the case where exactly one of them is a CS. Thanks to the long optimization setting, Ho and He reach good performance, low-rank versions of Ho (Ho\_2 and Ho\_3) give poorer results than classic Ho, which is not surprising. The parameterization specific to EHH also does not seem to improve the performances of the hypersphere. In all cases, LVGP delivers strong results.  In Section \ref{sec:experiments2}, we provide an analysis for datasets with no known groups which confirms these observations.

\section{Experiments with no group structure}
\label{sec:experiments2}

In our first experiments with a known group structure, we have clearly illustrated that nested kernels are superior to all other methods by a large margin. In the test cases without such structure, which we would like to investigate now, they are unfortunately inapplicable. We first discuss a new and computationally cheap approach to estimate groups from such datasets, based on target encoding. Importantly, our approach can be applied to cases where it is suspected that there is a group structure but is is unknown, but also to cases where there may be no underlying subdivision. We then perform the same experimental protocol as in the previous section, but this time on datasets without a group structure, and where we apply the nested kernel methods with the group structure obtained through our estimation procedure.

\subsection{Inferring the group structure from target encodings of the levels}
\label{sec:inferring_group_structure}

In this section, we investigate how to define groups of categorical variables in order to use the nested kernels when there is no prior knowledge about them. We suppose we have access to a single categorical variable (if several categorical variables are present, groups can be selected for each dimension separately). \citet{roustant2020group} suggest to train first a Gaussian process with another categorical kernel $\mathbf{T}^{\textbf{prox}}$. Then, given a number of clusters between $2$ and $C-1$, a clustering algorithm can be applied with the following distance:
\begin{equation}
d(z,z') = \left(\mathbf{T}^{\textbf{prox}}_{z,z} + \mathbf{T}^\textbf{prox}_{z',z'} - 2 \mathbf{T}^\textbf{prox}_{z,z'} \right)^{1/2}, \label{eq:proxdist}
\end{equation}
that is the pseudo-distance induced by the kernel associated to $\mathbf{T}^{\textbf{prox}}$ \citep{phillips2011gentle}. In practice, they propose to use hierarchical clustering, while the proxy categorical kernel can be chosen as a low-rank kernel such as LVGP. Unfortunately, the impact of this initialization is neither discussed nor illustrated numerically in \cite{roustant2020group}. In fact, $\mathbf{T}^{\textbf{prox}}$ may already be sufficient to reach the desired precision, so there is no need to apply the nested kernel method. On the other hand, if it has a poor predictive performance, we expect a misrepresentation of the distances between levels that do not show clear groups, and the final nested kernel obtained after clustering may be not adapted to the task. A second question concerns the number of groups. It is suggested to choose it as a parameter to be selected from an exhaustive grid search, for example with cross-validation. In addition to requiring several potentially costly training phases, it is common to handle small training sets, which limits cross-validation capabilities. In the following, we present a new approach that consists in representing categorical variables through their target encoding.

\begin{figure}
\begin{center}
\includegraphics[width=0.6\linewidth]{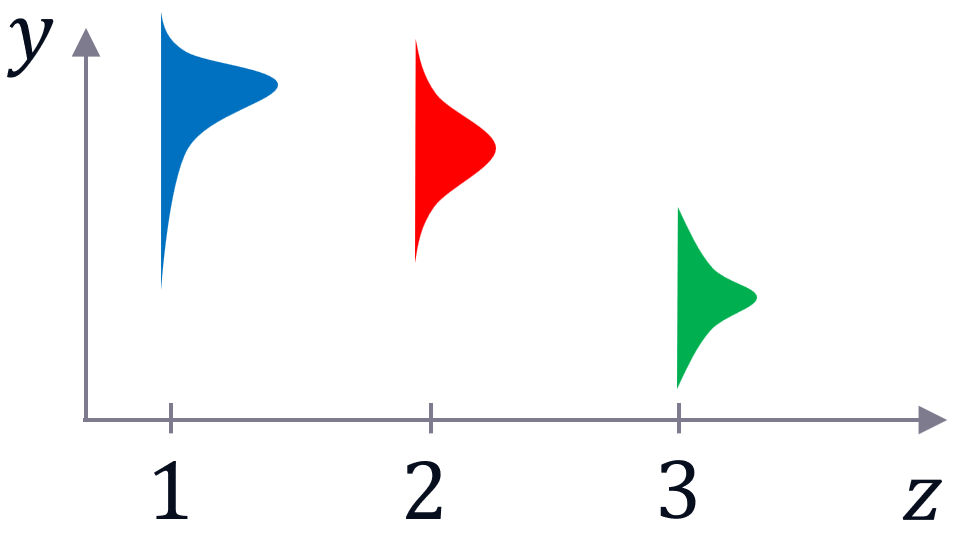}
\end{center}
\caption{Target encoding illustration. The z-axis represents 3 levels, and for each of them the conditional distribution of the output is represented on the y-axis. Intuitively, levels 1 and 2 should be grouped together since their target encoding distributions are close.  \label{fig:target_encodings}}
\end{figure}

\paragraph{Target encoding of categorical variables}~\\
Recall that we consider a training set with categorical inputs $z^{(1)}, \cdots, z^{(N)} \in \llbracket C \rrbracket$ and output scalars $y^{(1)}, \cdots, y^{(N)} \in \mathbb{R}$. For a level $c=1\cdots, C$, we consider the distribution of the output $y$ conditioned on the input variable $z$ being equal to $c$. This distribution can be approximated by the empirical distribution of the observations, i.e. $\nu_c := \frac{1}{|\{i : z^{(i)}=c\}|} \sum\limits_{i : z^{(i)}=c} \delta_{y^{(i)}}$. Such representation is called a target encoding of the level, since it contains a quantitative feature summary of the level obtained through the output values (the target). We can thus transfer the problem of defining a distance between levels as in (\ref{eq:proxdist}) to the much more manageable problem of computing a dissimilarity between the encodings. In particular, for empirical distributions, we can use any statistical divergence, including the Wasserstein distance \citep{computational_ot} or the Maximum Mean Discrepancy \citep{mmd1}, see for example \citet{da2025distributional}. Intuitively, we measure the proximity of levels by their impact on the output. Figure \ref{fig:target_encodings} gives an idealized illustration of target encoding.

 In practice however, using classical divergences may not be sufficient when the support of the empirical distributions is small. For most datasets, we work with less than a dozen of samples by level, meaning that estimation of divergences will therefore not be relevant. We instead choose to use simple features of the distributions, namely their mean and standard deviation. The level $c$ is thus embedded in $\mathbb{R}\times\mathbb{R}_+$ as $\psi(c) := (\mu_c, \sigma_c) $ where $\mu_c := \frac{1}{|\{i : z^{(i)}=c\}|} \sum\limits_{i : z^{(i)}=c} y^{(i)}$ and $\sigma_c := \sqrt{\frac{1}{|\{i : z^{(i)}=c\}|}  \sum\limits_{i : z^{(i)}=c} (y^{(i)}-\mu_c)^2}$. We call this representation target MSD (Mean and Standard Deviation). Note the similarity with a two-dimensional representation from  LVGP, but our proposal is weakly supervised and thus can be computed with almost no computational cost.

\paragraph{Clustering with automatic selection of the number of groups}~\\
Suppose now we have access to a distance matrix $\mathbf{D}^{\textbf{prox}} \in \mathbb{R}^{C\times C}$. It can either be a distance induced by a categorical kernel $\mathbf{T}^{\textbf{prox}}$ or a Euclidean distance matrix when levels are embedded in a Euclidean space such as with our proposed target encoding. The first step is to cluster the levels in $Q$ groups from their distances $\mathbf{D}^{\textbf{prox}}$: here we use agglomerative hierarchical clustering \citep{murtagh2012algorithms} but any other clustering algorithm may be considered. The remaining question is how to choose the number of groups $Q$. Instead of training a GP with a group kernel several times with different numbers of clusters $2\leq Q \leq C-1$, we suggest instead to use a heuristic. In our experiments, we use the mean Silhouette Coefficient \citep{rousseeuw1987silhouettes}, which measures how well each data point lies within its cluster, and how well separated the clusters are from each other. Other options are discussed in \ref{appendix:clustering}, and in particular how to handle $Q=1$ or $Q=C$ clusters.

\begin{figure}[h!]
\begin{center}
\includegraphics[width=0.19\linewidth]{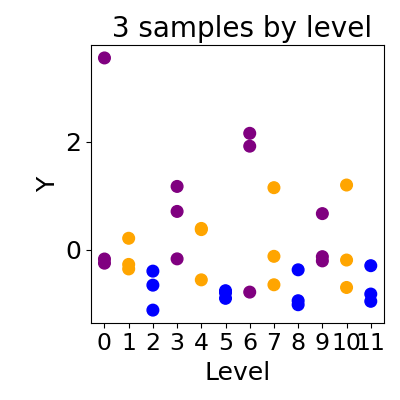}
\includegraphics[width=0.19\linewidth]{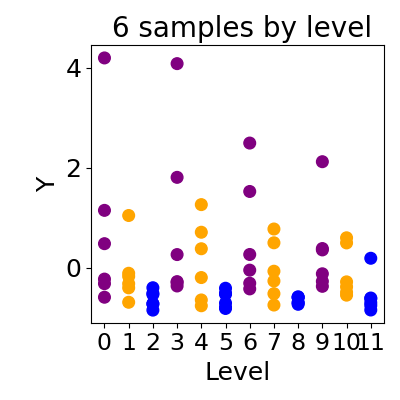}
\includegraphics[width=0.19\linewidth]{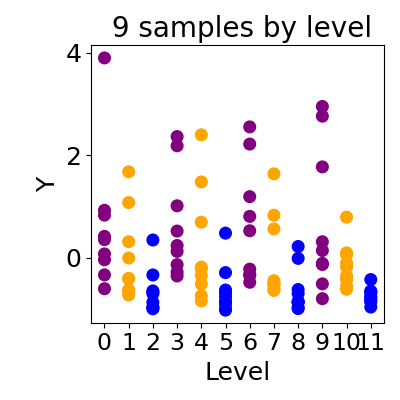}
\includegraphics[width=0.19\linewidth]{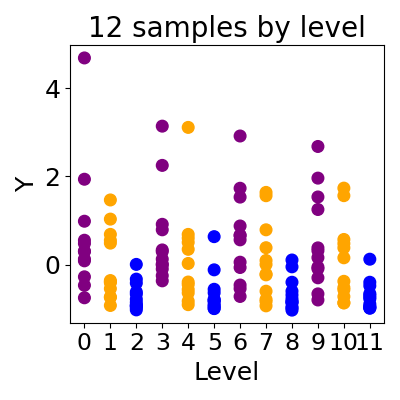}
\includegraphics[width=0.19\linewidth]{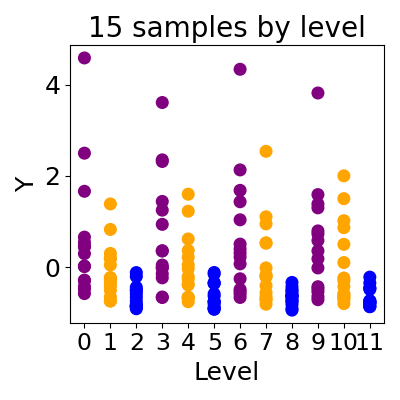}
\end{center}
\caption{Target distributions of the 12 levels of the beam bending problem. From left to right: 3/6/9/12/15 samples by level. Each color represents a different group. \label{fig:target_distribs_BB}}
\end{figure}

\paragraph{Visualization of the group initialization strategies.}~\\
For illustration, we compare initialization strategies using latent variables and target encoding. We consider the beam bending problem (see \ref{appendix:presentation_datasets}) with varying dataset size, and we always take the first experimental seed. We recall that there are 3 groups of levels corresponding to the different filling configurations of the shapes. We start by representing the empirical distributions of the outputs for each level in Figure \ref{fig:target_distribs_BB}.

\begin{figure}[H]
\begin{center}
\includegraphics[width=0.19\linewidth]{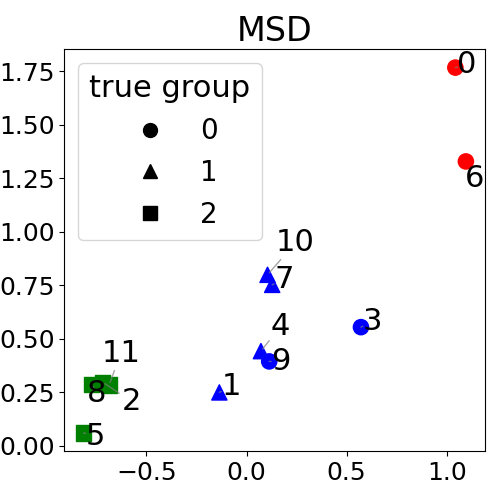}
\includegraphics[width=0.19\linewidth]{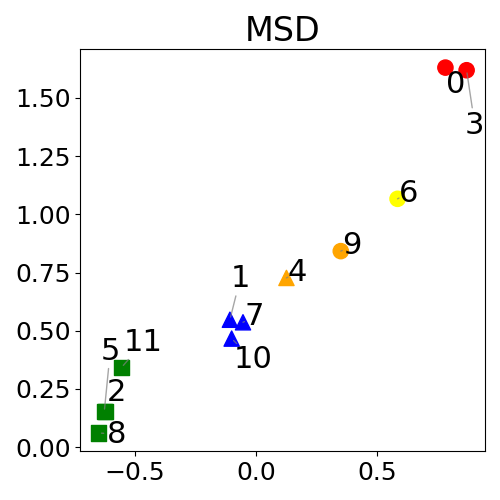}
\includegraphics[width=0.19\linewidth]{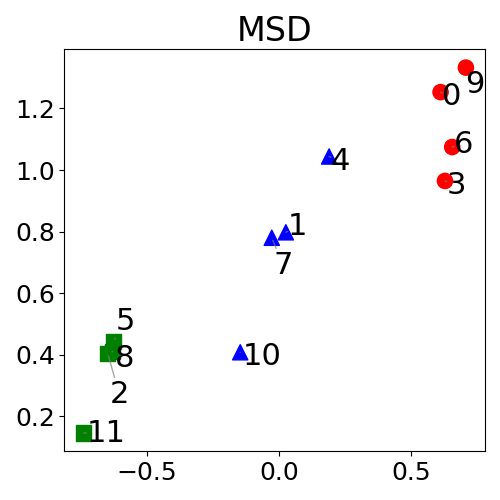}
\includegraphics[width=0.19\linewidth]{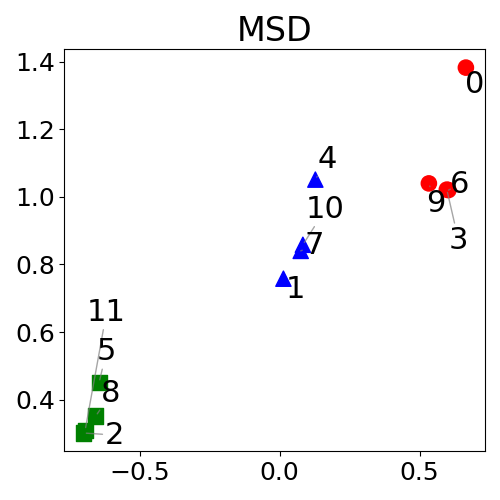}
\includegraphics[width=0.19\linewidth]{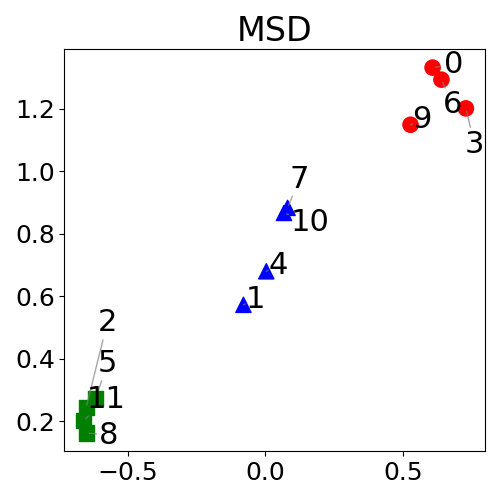}
\includegraphics[width=0.19\linewidth]{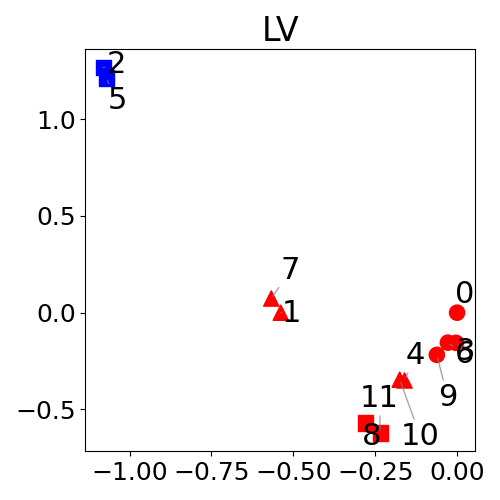}
\includegraphics[width=0.19\linewidth]{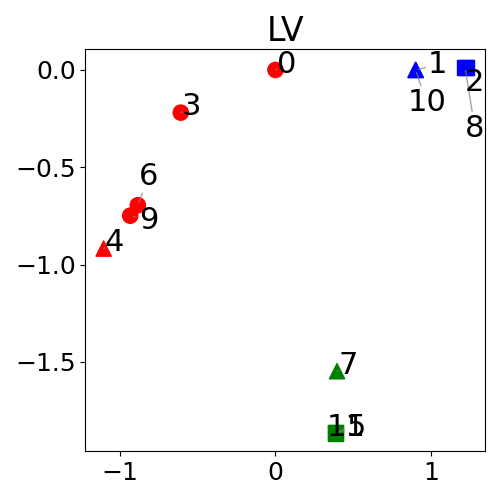}
\includegraphics[width=0.19\linewidth]{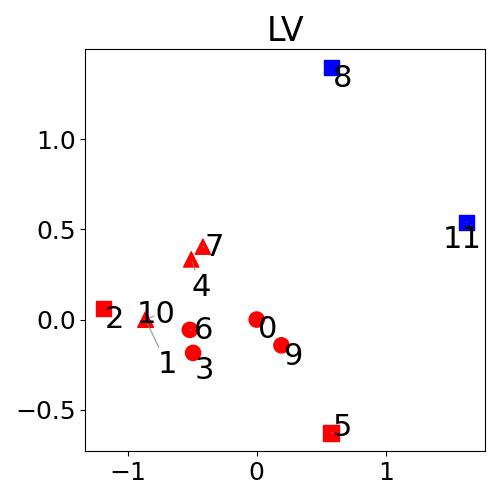}
\includegraphics[width=0.19\linewidth]{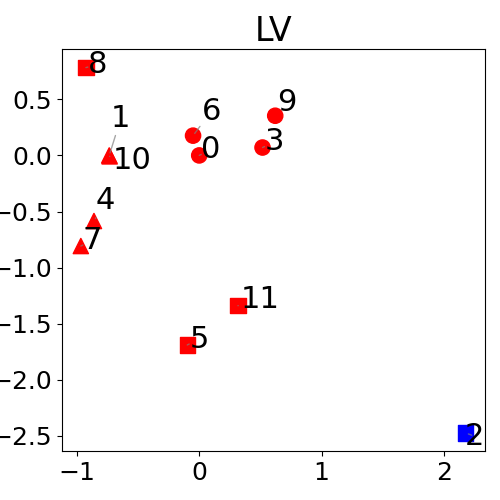}
\includegraphics[width=0.19\linewidth]{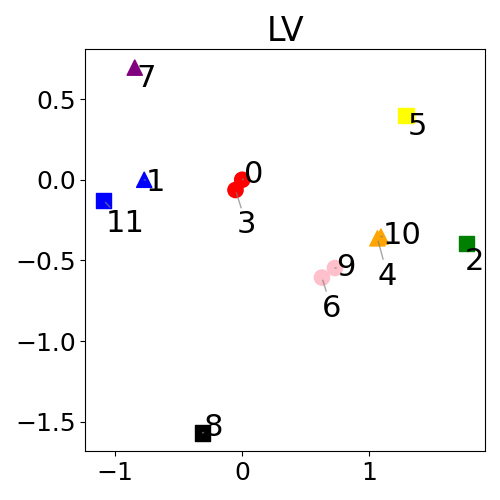}
\end{center}
\caption{Clustering of the levels based on their target encoding with mean and standard deviation (top) or their LVGP latent variables (bottom) for the beam bending dataset. From left to right: 3/6/9/12/15 samples by level. Colors represent clusters, and the marker styles give the true group. \label{fig:clusters}}
\end{figure}

We then compute the means and standard deviations of the conditional distributions to embed each level in a 2D space. We compare such embeddings with the latent variables obtained after training LVGP in Figure \ref{fig:clusters}, and also show the obtained clusters after running the clustering procedure introduced before. We first note that LVGP fails to identify the true groups regardless of the number of samples per level. For target encoding, the clustering quality increases with the number of samples per level. From 9 samples by level and onward, it can retrieve the true groups successfully. For 3 and 6 samples by level, the group $\{2,5,8,11\}$  is well identified, the group $\{1,4,7,10\}$ is partially identified, and the group $\{0,6,3,9\}$ is harder to detect. Looking back at the representations of the target distributions in Figure \ref{fig:target_distribs_BB}, this is not surprising: with few samples per level, groups are less obvious to identify.

Finally, if we investigate again our previous numerical experiment on the beam bending dataset, but this time with a nested kernel where the groups are estimated by LVGP or target encoding, we see that a good cluster initialization can help improve scores for this experiment, as illustrated in Table \ref{tab:LV_vs_MSD_scores}. As expected from the clustering quality discussed above, target encoding with MSD is able to reach the same accuracy as if we knew the groups beforehand. Extensive comparisons on other datasets with known groups between the original nested kernels and the ones where the groups are estimated can be found in \ref{appendix:clustering}.

\begin{table*}[h!]
\setlength{\tabcolsep}{6pt}
\caption{Comparison of the RRMSE for the Nested\_He\_He kernel on one experiment of the beam bending problem. Each line corresponds to a group selection strategy.}
\label{tab:LV_vs_MSD_scores}
\centering
\begin{tabular}{|cccccc|} 
\toprule
Samples by level & 3 & 6 & 9 & 12 & 15 \\ 
\midrule
True & $9.81 \times 10^{-2}$ & $2.04 \times 10^{-2}$ & $8.73 \times 10^{-3}$ & $6.79 \times 10^{-3}$ & $2.04 \times 10^{-3}$ \\ 
MSD & $2.00 \times 10^{-1}$ & $1.70 \times 10^{-2}$ & $8.73 \times 10^{-3}$ & $6.79 \times 10^{-3}$ & $2.04 \times 10^{-3}$ \\ 
LV & $2.67 \times 10^{-1}$ & $8.16 \times 10^{-2}$ & $1.31 \times 10^{-2}$ & $8.54 \times 10^{-3}$ & $1.7 \times 10^{-2}$ \\  
 \bottomrule
\end{tabular}
\end{table*}

In what follows, for datasets with no group structure, we will then use these estimated nested kernels instead of the original ones, see Table \ref{tab:kernels2} for a summary.

\begin{table}[ht!]
\caption{List of kernels for experiments with no group structure only.}
\small
\centering
\begin{tabular}{|llc|}
\toprule
\textbf{Category} & \textbf{Name} & \textbf{Description} \\
\midrule
\multirow{2}{*}{\shortstack{Nested Kernels \\ Estimated groups}}
  & Nested\_[BETW]\_[WITH]\_LV & LVGP to initialize groups  \\
  & Nested\_[BETW]\_[WITH]\_MSD & Target mean and sd to initialize groups  \\
\bottomrule
\end{tabular}
\label{tab:kernels2}
\end{table}

\subsection{Presentation of the datasets}

Table \ref{tab:comparison2} presents the datasets with no group structure used in the study. Like in the previous experiments, for all datasets, designs of experiments of different sizes are generated with 50 independent replications. The datasets we select cover a wide range of situations: some have 1 or 2 categorical variables with numbers of levels ranging from 3 to 24, and with different training sizes.

\begin{table*}[h]
\setlength{\tabcolsep}{6pt}
\caption{Description of all datasets with no group structure: number of continuous and categorical variables (number of levels in parenthesis) and size of training and test datasets. Groups=False means that the groups are unknown.}
\label{tab:comparison2}
\centering
\begin{tabular}{|ccccccc|} 
 \toprule
Name & Cont & Cat & Train & Test & Groups & Source\\
 \midrule
 Borehole & 6 & 2 (3-4) & 36/72/108/144/180 & 1008 & False & \cite{zhang2020latent}\\
 Borehole2 & 6 & 1 (12) & 36/72/108/144/180 & 1008 & False & \cite{zhang2020latent}\\
 OTL & 4 & 2 (4-6) & 72/144/216 & 1008 & False & \cite{zhang2020latent}\\
 OTL2 & 4 & 1 (24) & 72/144/216 & 1008 & False & \cite{zhang2020latent}\\
 Piston & 5 & 2 (5-3) & 45/90/135 & 1005 & False & \cite{zhang2020latent}\\
 Piston2 & 5 & 1 (15) & 45/90/135 & 1005 & False & \cite{zhang2020latent}\\
 Goldstein & 2 & 1 (9) & 27/54/81/108/135 & 999 & False & \cite{pelamatti2021mixed}\\
 \bottomrule
\end{tabular}
\end{table*}

\subsection{Comparison of the methods}

As before, we use the RRMSE as an evaluation metric and compare methods on individual datasets and through performance profiles. In addition, we also propose a new metric that balances accuracy and running time, since this may be of interest in practice to choose a method.

\begin{figure}[H]
\begin{center}
\includegraphics[width=0.49\linewidth]{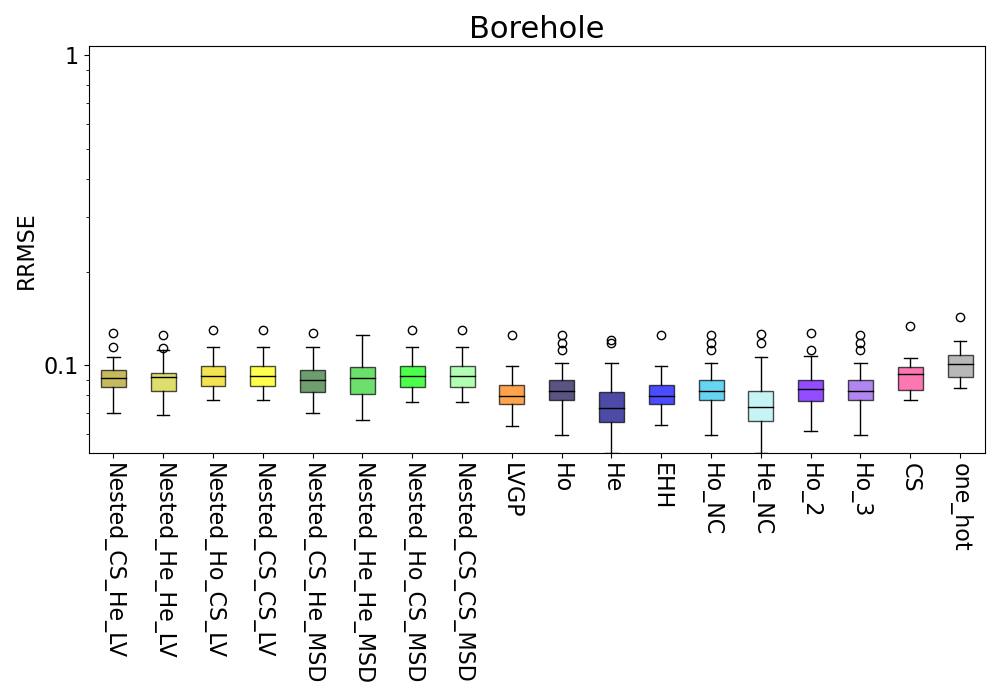}
\includegraphics[width=0.49\linewidth]{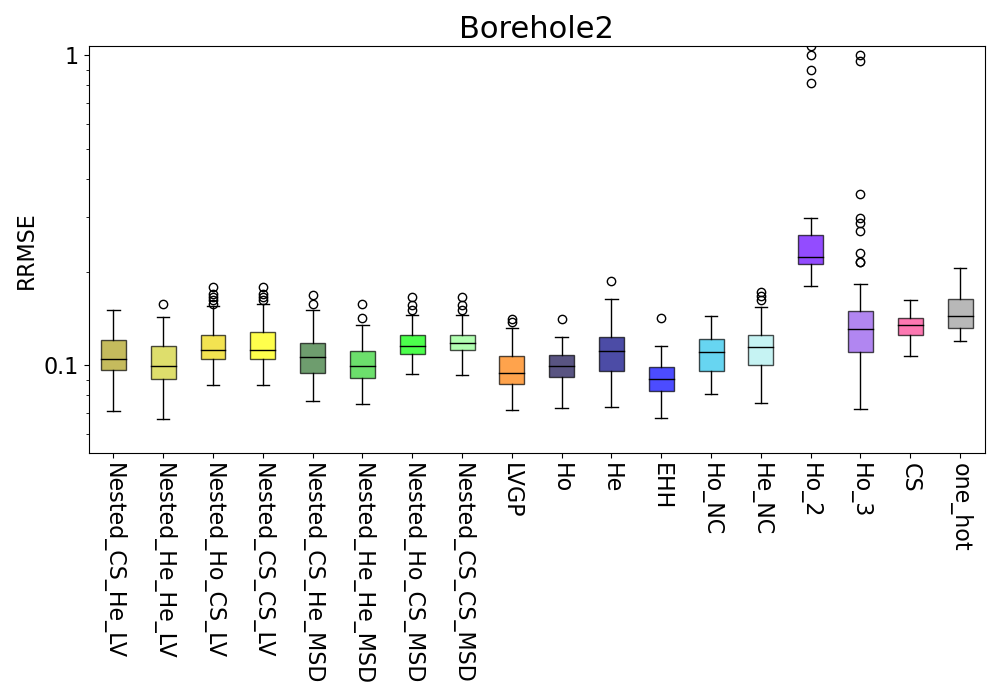}
\includegraphics[width=0.49\linewidth]{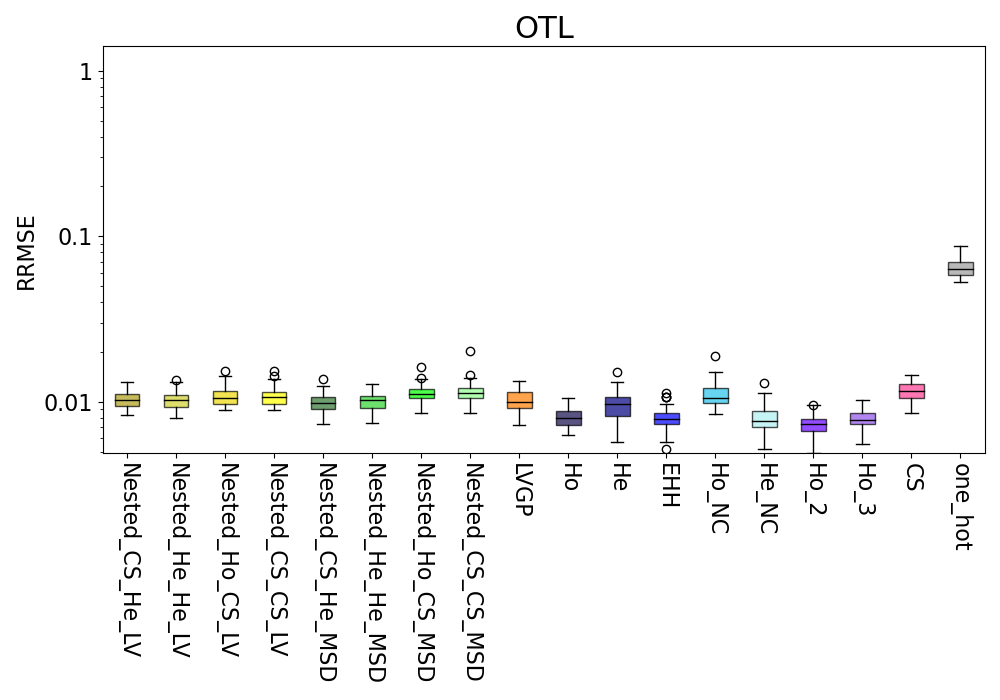}
\includegraphics[width=0.49\linewidth]{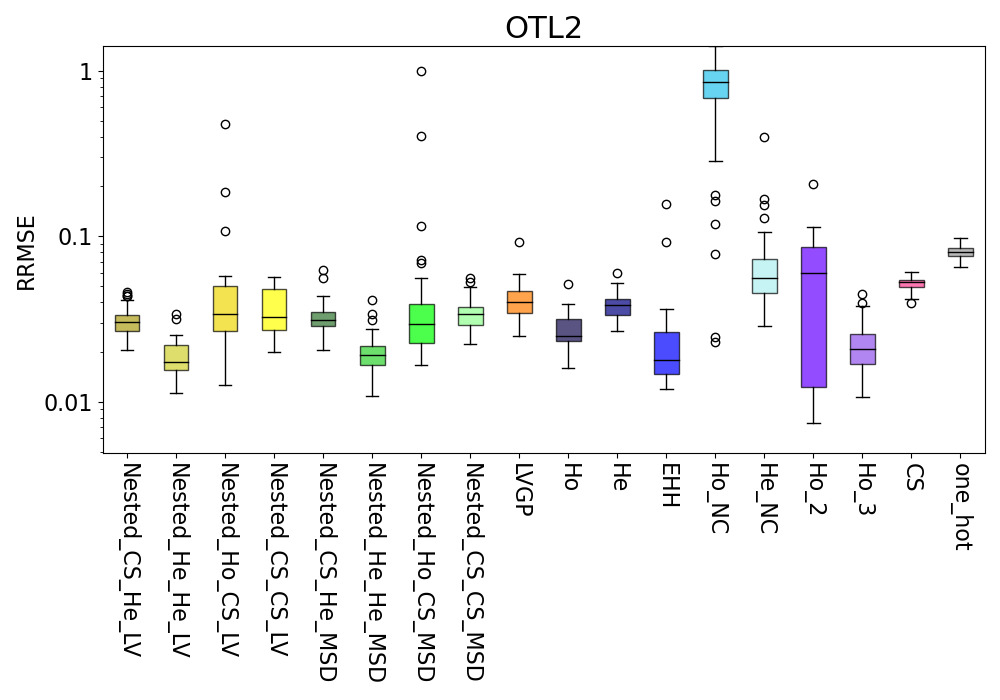}
\includegraphics[width=0.49\linewidth]{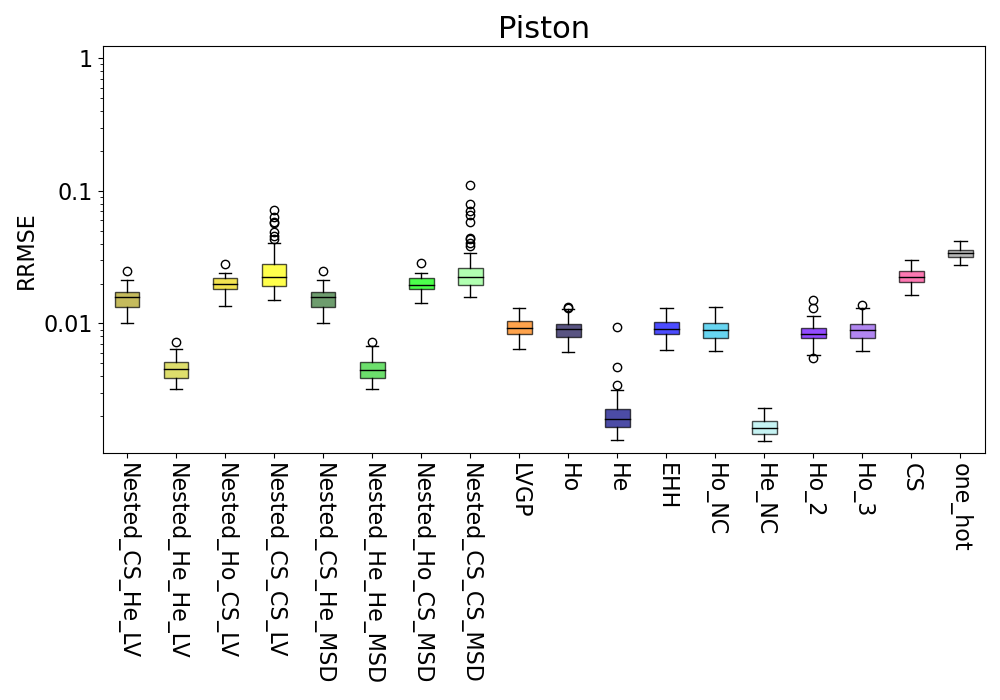}
\includegraphics[width=0.49\linewidth]{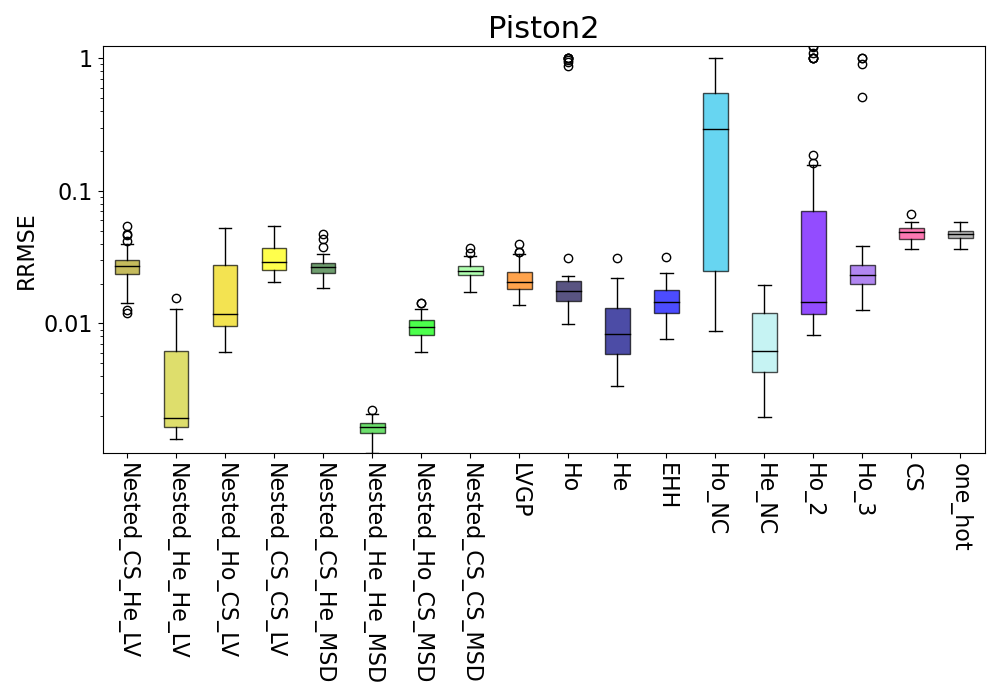}
\end{center}
\caption{OTL, Piston and Borehole dataset: categorical variables separated (left) and fused (right). Boxplots of the RRSME over the 50 experiments for all methods, "long" optimization setting.}\label{fig:boxplots_shared_scale}
\end{figure}

\paragraph{Results on each dataset}~\\
\label{sec:results_datasets2}
Results provided in the repository containing the code show that on average the nested kernels still perform better than competitors. Here, we will only comment on specific experiments that we think are of practical interest. Looking closely at the dataset list in Table \ref{tab:comparison2}, one can observe that we have considered variants of the original OTL, Piston and Borehole dataset, where categorical inputs are either split (original test case) or fused into a single new categorical input with more levels (variant). We proposed to investigate these variants because, in practice, one may wonder if merging all categorical inputs may be beneficial for training a GP. Figure \ref{fig:boxplots_shared_scale} shows that, for the three datasets, it seems preferable to consider the categorical variables separately rather than merging them. The difference is more pronounced for the OTL dataset. Performance is also generally most degraded for hypersphere parameterizations, which suffer from an increase in the number of levels. 

\begin{figure}[H]
    \centering
    \begin{subfigure}{0.9\textwidth}
        \centering
        \includegraphics[width=1.0\linewidth]{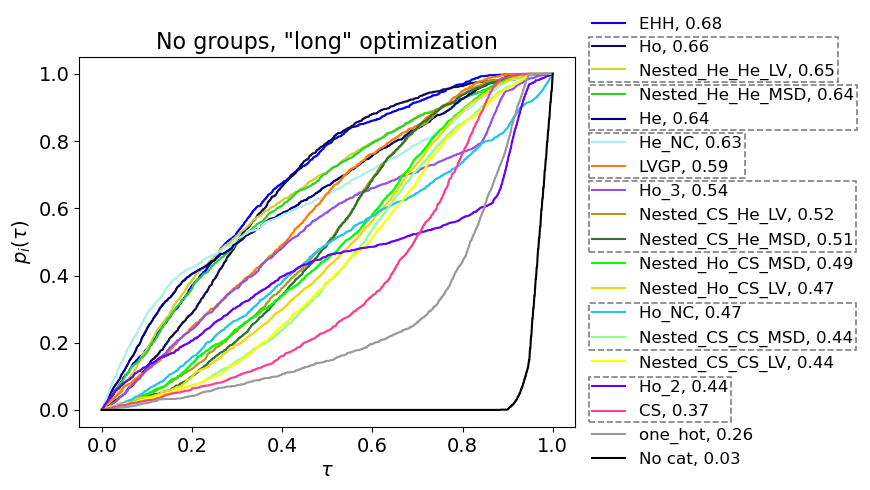}
        \caption{"long" optimization}
        \label{fig:profile_all_RRMSE_noGK_LONG}
    \end{subfigure}
    \hfill
    \begin{subfigure}{0.9\textwidth}
        \centering
        \includegraphics[width=1.0\linewidth]{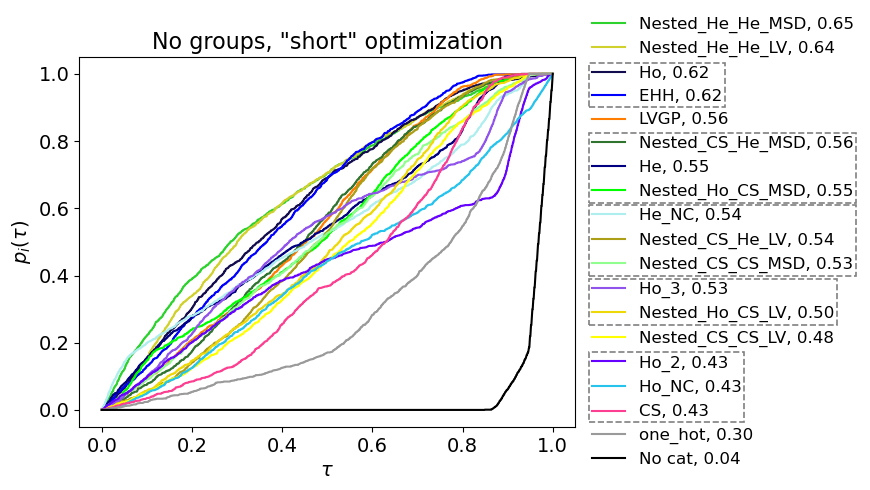}
        \caption{"short" optimization}
        \label{fig:profile_all_RRMSE_noGK_SHORT}
    \end{subfigure}
    \caption{Performance profiles using the RRMSE, "long" (top) and "short" (bottom) optimization setting. Only datasets with no groups are considered. The score after the name of the method $i$ is $\textbf{AUC}(p_i)$, methods are sorted in AUC decreasing order.}
    \label{fig:profile_all_RRMSE_noGK}
\end{figure}

\paragraph{Performance profiles}~\\
We now compute the performance profiles for all methods, they are given in Figure \ref{fig:profile_all_RRMSE_noGK_LONG} for the long optimization setting and in Figure \ref{fig:profile_all_RRMSE_noGK_SHORT} for the short optimization setting.

In contrast to the results from Section \ref{sec:experiments1}, EHH, Ho, He and He\_NC yield remarkably strong performance. This finding differs from initial expectations, as hypersphere models are often ignored in comparative studies due to lower-quality results in previous studies. Based solely on these indicators, we should conclude that hyperspheres are preferable to datasets without known groups. However, if we change the optimization options, we arrive at a very different conclusion. In Figure \ref{fig:profile_all_RRMSE_noGK_SHORT} we use the "short" optimization instead of the "long" optimization setting. Remember that the "short" optimization relies on the default options from scipy's L-BFGS-B algorithm as we detailed in Section \ref{sec:implementation_details}. With these parameters, Nested\_He\_He with automatic selections via LVGP or target encodings outperforms other models on datasets with unknown groups. As discussed in more detail in \ref{appendix:optimization}, this difference in rankings between "long" and "short" optimizations can be explained by a significant deterioration in the scores of hypersphere models. It is therefore possible to obtain very good scores for hyperspheres for these datasets, but this takes hours instead of a few minutes of training, which is unfortunately not feasible for standard use cases. When using default optimization options and thus manageable run times, we realize that nested kernels offer the most robust alternative. 

We find that group initializations with LVGP or MSD are nearly equivalent. However, unlike in the case of known groups, the choice of between and within covariance structures is more pronounced. For hypersphere models, Ho\_NC and low rank variants are once again inferior to other full-rank variants. Nevertheless, EHH parameterization seems to be more useful on these datasets. Lastly, while LVGP is still strong, it appears clearly that strategies based on one-hot encoding or CS should be avoided because they show degraded performances. 

Finally, we also provide in Figure \ref{fig:profile_all_RRMSE_FULL} from \ref{appendix:performance_profiles} the performance profiles when we include all datasets (with or without groups). It can be seen that Nested\_He\_He\_LV and Nested\_He\_He\_MSD achieve the best overall performance, even in the case of "long" optimization.

\begin{figure}[H]
\begin{center}
\includegraphics[width=0.98\linewidth]{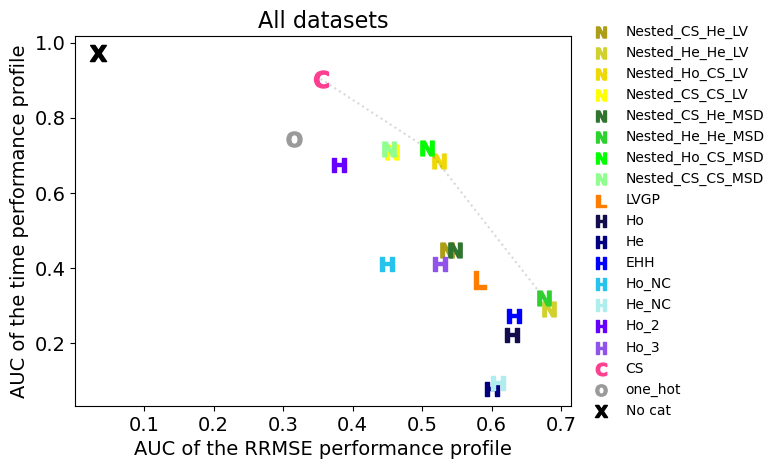}
\end{center}
\caption{AUC based on sequential time versus AUC based on RRMSE, "long" optimization setting. The dotted grey line represents the "Pareto front". \label{fig:rank_time_vs_rrmse}}
\end{figure}

\paragraph{Tradeoff between computation time and RRMSE}~\\
In previous representations, only the RRMSE was taken into account. However, computation time is also an indicator that can influence the choice of the kernel in practical studies, as was already touched upon with both optimization settings. 
We now also investigate time-based indicators by evaluating the sequential time required to train the GP model. As in the case of RRMSE, performance profiles can be defined for methods based on running times across all experiments. Computation times for nested methods with automatic group initializations (LV and MSD) do not include the identification of groups (this step is very fast thanks to the target encoding of the MSD representation contrary to LV). Importantly, for this paragraph and study only, we consider all possible datasets, and the methods available on each one of them. In Figure \ref{fig:rank_time_vs_rrmse}, each method is represented by a point, with the x-axis corresponding to the AUC of its performance profile based on RRMSE, and the y-axis to the AUC of its time-based performance profile. A good method is therefore located as far up on the right as possible. Observe first that rankings based on computation times strongly correlate with the numbers of parameters depicted in Figure \ref{fig:chronology}.\\

One-hot encoding and compound symmetry have low computation times, but they have the worst prediction capabilities. On the contrary, hypersphere models have the highest computation time, and average prediction capabilities. Using low-rank hypersphere variants naturally lowers the computation time, but it decreases the quality of predictions. Nested kernels having both a hypersphere as within and between correlation have times similar to hyperspheres, but they offer better predictive performance. The computation times of LVGP are rather high, but it is still a reliable approach that is independent on prior group structure. We also represent in Figure \ref{fig:rank_time_vs_rrmse} a "Pareto front", which indicates methods that achieve the best trade-offs between time and learning capacity. Note that this is mainly for illustration purposes, as the lines have no formal definition. This suggests to use nested kernels with automatic groups and heteroscedastic models as between and within covariances for the best learning capacity, but at the cost of a higher computation time. One may replace the between covariances with a homoscedastic structure to reduce the complexity, while keeping average performances. Finally, the CS model seems to offer the best performance when time is even more limited.

\section{Conclusion}
\label{sec:conc}

In this paper, we detail an extensive comparative study of categorical kernels used for Gaussian process regression in mixed spaced (with continuous and categorical input variables). 23 categorical kernels are evaluated on 42 datasets widely used in the literature, with two optimization settings. To our knowledge, this is the first time such an advanced comparison of the different kernels found in the literature is carried out, with a ranking of methods based on new performance metrics on several dataset experiments. In addition, our study is entirely reproducible with the accompanying code.\\

We also propose an innovative way of defining groups of levels when they are not known or when there is no prior information on the group structure, using clustering based on a target encoding representation of the levels. This new representation delivers performance comparable to that based on LVGP initialization, but without the need to pre-train another model. It is clear from our study that nested kernels strategies outperform other kernels in terms of model error when groups exist and are known. Even in the absence of known groups, nested variants with automatic group selection using LVGP or target encoding are once again among the best methods. However, the choice of within and between correlation types in nested kernels matters, He/He being the best choice but at the cost of larger computing times. Contrary to the popular belief, hypersphere models are competitive from the perspective of performances, but their high computation time is not cost-effective when compared to other methods. They present good performances only when the hyperparameters of the optimization routine (tolerance, maximal number of evaluations) are chosen beyond their default values. Low-rank variants of hypersphere do not appear to be a reliable alternative, and also suffer from instabilities, in particular with negative correlations, but other parameterization choices like EHH can somewhat improve the performances. CS and one-hot encoding should not be used if good performance is required, while LVGP is always among the best methods.  

\section*{Acknowledgements}
This work was partially supported by the Agence Nationale de la Recherche through the SAMOURAI (Simulation Analytics and Metamodel-based solutions for Optimization, Uncertainty and Reliability AnalysIs) project under grant ANR20-CE46-0013 and the EXAMA (Methods and Algorithms at Exascale) project under grant ANR-22-EXNU-0002.









\clearpage
\begin{appendix}
\renewcommand\thefigure{\arabic{figure}}
\renewcommand\thetable{\arabic{table}}
\renewcommand\thesubsection{\arabic{subsection}}
\renewcommand\theequation{\arabic{equation}}

\section{Presentation of the datasets}
\label{appendix:presentation_datasets}

All training and test datasets except Goldstein are generated in the same way: for categorical variables, the same number of samples is used for each level (or tuple of levels in the multivariate case), and for continuous variables we create a Sliced Latin Hypercube Design (SLHD) \citep{qian2012sliced}. 50 independent replications are created to form the training sets for a given dataset. Remark that the designs are not optimized using e.g. a maximin LHS \citep{stein1987large} since we want to keep some variability between experiments. For Goldstein, since the continuous variables are subject to constraints, we generate points by a rejection procedure combined with a uniform distribution. In practice, a standard scaling is applied to all outputs as well as all input continuous variables. For the datasets having two categorical variables (OTL, Piston and Beam Bending), we also consider variants of the dataset where these levels are merged to form a single categorical variable.

\begin{figure}[h!]
\begin{center}
\includegraphics[width=0.6\linewidth]{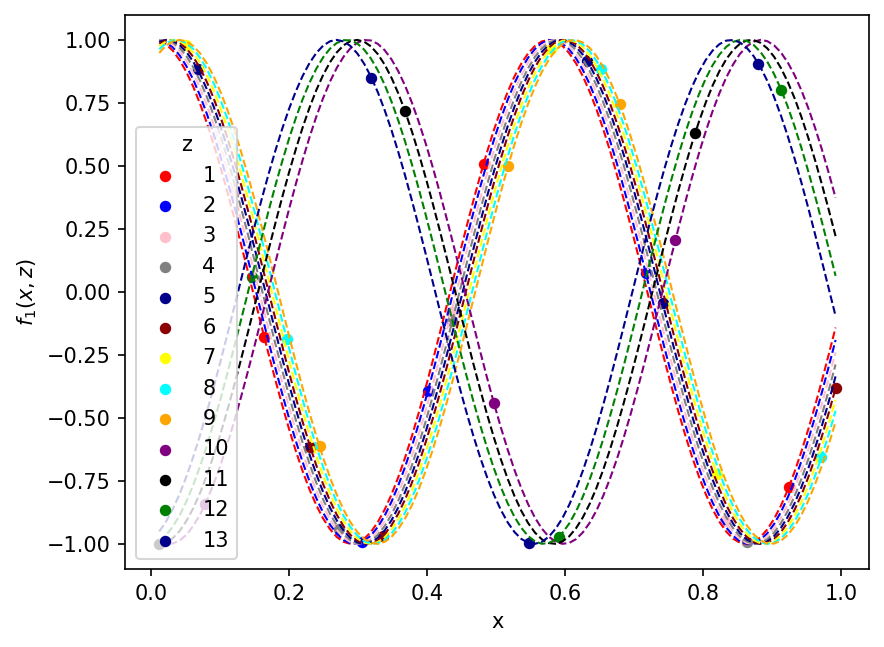}
\end{center}
\caption{$f_1$. Bullet points represent training samples, and lines represent the true functions for each level. \label{fig:f1_GK_13}}
\end{figure}

\paragraph{Analytical function $f_1$}~\\
The function is defined for a continuous variable and a categorical input with 13 levels as follows:
\begin{equation*}
 f_1(x,z) = \cos(7 \pi \frac{x}{2} + (0.4 + \frac{z}{15})\mathbb{1}_{\{z>9\}} - \frac{z}{20})
\end{equation*}
where $x\in [0,1]$, $z\in \llbracket 13 \rrbracket$. There are two groups of levels ($\{1,2,3,4,5,6,7,8,9\}$ and $\{10,11,12,13\}$) that appear clearly in Figure \ref{fig:f1_GK_13}. We consider 5 different sizes for the training sets 39/78/117/156/195 corresponding to 3/6/9/12/15 samples by level, respectively.

\begin{figure}[h!]
\begin{center}
\includegraphics[width=0.6\linewidth]{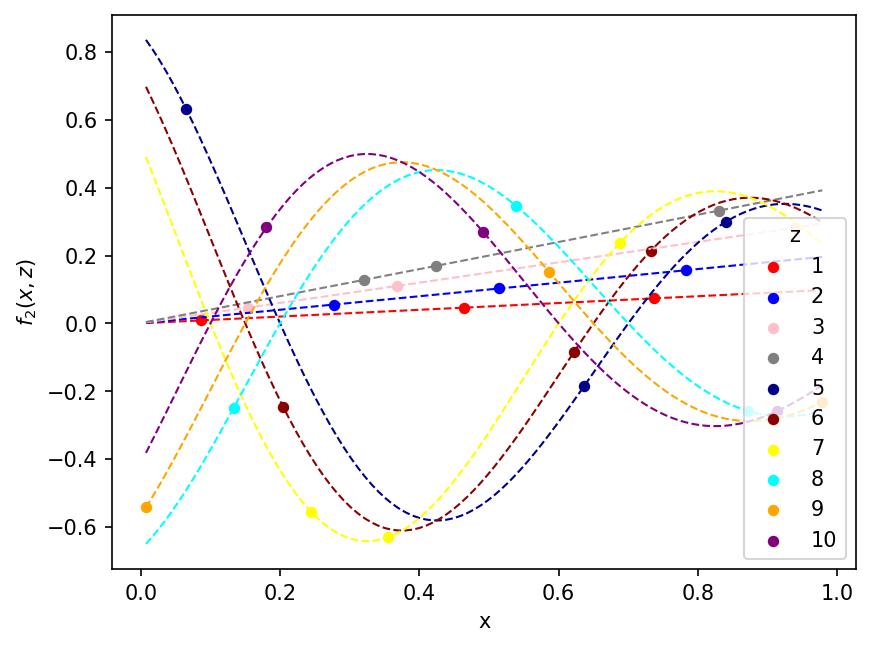}
\end{center}
\caption{$f_2$. Bullet points represent training samples, and lines represent the true functions for each level. \label{fig:f2_GK_10}}
\end{figure}

\paragraph{Analytical function $f_2$}~\\
The function is defined for a continuous variable and a categorical input with 10 levels as follows:
\begin{equation*}
  f_2(x,z) =
    \begin{cases}
      x + 0.01(x-\frac{1}{2})^2\frac{z}{10} & \text{if $z=1,2,3,4$}\\
      0.9\cos(2\pi (x+(z-4)\frac{z}{20}))e^{-x} & \text{if $z=5,6,7$}\\
      -0.7\cos(2\pi (x+(z-7)\frac{z}{20}))e^{-x} & \text{if $z=8,9,10$}
    \end{cases} 
\end{equation*}
where $x\in [0,1]$, $z\in \llbracket 10 \rrbracket$. There are three explicit groups of levels ($\{1,2,3,4\}$, $\{5,6,7\}$ and $\{8,9,10\}$) that appear clearly in Figure \ref{fig:f2_GK_10}. We consider 5 different sizes for the training sets 30/60/90/120/150 corresponding to 3/6/9/12/15 samples by level, respectively.

\paragraph{Borehole}~\\
The continuous Borehole function is:
\begin{equation*}
f_B(r, T_u, H_u, T_l, L, K_w) = 2\pi T_u (H_u - H_l) \left\{ \ln(\frac{r}{r_w}) \left(1+2 \frac{LT_u}{\ln(\frac{r}{r_w})r_w^2 K_w}+ \frac{T_u}{T_l}\right) \right\}^{-1}
\end{equation*}
where $r\in [100,50000]$, $r_w\in [0.05,0.15]$, $T_u\in [63070,115000]$, $H_u\in [990,1110]$, $T_l \in [63.1,116]$, $H_l \in [700,820]$, $L\in [1120,1680]$, $K_w \in [9855,12045]$, see \cite{morris1993bayesian} for a more precise description
of the variables. $r_w$ and $H_l$ are treated as categorical variables with respectively 3 and 4 levels whose associated continuous values are evenly distributed across their definition spaces. This gives a total of 6 continuous and 2 categorical variables. The Borehole2 function in Table \ref{tab:comparison2} corresponds to the same dataset, but the two categorical variables are fused into a single one with $12$ levels.  We consider 5 different sizes for the training sets 36/72/108/144/180 corresponding to 3/6/9/12/15 samples by level tuple, respectively.

\paragraph{OTL}~\\
The continuous OTL function is:
\begin{equation*}
f_O(R_{b1}, R_{b2}, R_f,R_{c1}, R_{c2}, \beta) = \frac{(V_{b1}+0.74)\beta (R_{c2}+9)}{\beta(R_{c2}+9)+R_f}+\frac{11.35 R_f}{\beta(R_{c2}+9)+R_f} + \frac{0.74 R_f \beta (R_{c2}+9)}{R_{c1}(\beta(R_{c2}+9)+R_f)}
\end{equation*}
where $V_{b1}=\frac{12 R_{b2}}{R_{b1}+R_{b2}}$, $R_{b1}\in [50,150]$, $R_{b2}\in [25,70]$, $R_f\in [0.5,3]$, $R_{c1}\in [1.2,2.5]$, $R_{c2}\in [0.25, 1.2]$, $\beta\in [50,300]$, see \cite{ben2007modeling} for a more precise description of the variables. $R_f$ and $\beta$ are treated as categorical variables with 4 and 6 levels whose associated continuous values are respectively $(0.5,1.2,2.1,2.9)$ and $(50,100,150,200,250,300)$. This gives a total of 4 continuous and 2 categorical variables. The OTL2 function in Table \ref{tab:comparison2} corresponds to the same dataset, but the two categorical variables are fused into a single one with $24$ levels.  We consider 3 different sizes for the training sets 72/144/216 corresponding to 3/6/9 samples by level tuple, respectively.

\paragraph{Piston}~\\
The continuous Piston function is:
\begin{equation*}
f_P(M,S,V_0,k,P_0,T_a, T_0) = 2\pi \sqrt{M^{-1}\left( k+ S^2 \frac{P_0 V_0 T_\alpha}{T_0} \left\{\frac{S}{2k}\left( \sqrt{A^2 +4k\frac{P_0 V_0 T_\alpha}{T_0}}-A\right) \right\}^{-2}\right)}
\end{equation*}
where $A=P_0 S + 19.62M-\frac{kV_0}{S}$, $M\in [30,60]$, $S\in [0.005,0.02]$, $V_0\in [0.002,0.01]$, $k\in [1000,5000]$, $P_0\in [90000,110000]$, $T_a\in [290,296]$, $T_0\in [340,360]$, see \cite{sack1989design} for a more precise description of the variables. $P_0$ and $k$ are treated as categorical variables with respectively 3 and 5 levels whose associated continuous values are evenly distributed across their definition spaces. This gives a total of 5 continuous and 2 categorical variables. The Piston2 function in Table \ref{tab:comparison2} corresponds to the same dataset, but the two categorical variables are fused into a single one with $15$ levels.  We consider 3 different sizes for the training sets 45/90/135 corresponding to 3/6/9 samples by level tuple, respectively.

\paragraph{Beam bending}~\\\label{paragraph_beam_bending}
\begin{figure}[h!]
\begin{center}
\includegraphics[width=0.6\linewidth]{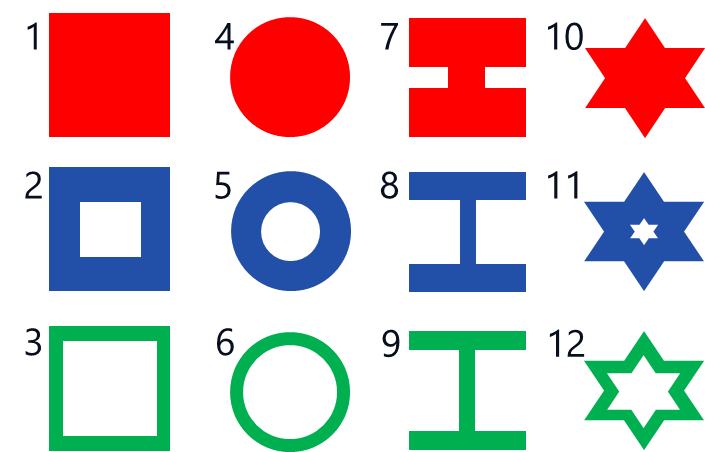}
\end{center}
\caption{The twelve shapes of the beam bending problem. The three groups can be identified thanks to the colors. Figure inspired by \cite{roustant2020group}. \label{fig:beam_bending_shapes}}
\end{figure}
We consider the Euler-Bernoulli beam bending problem \citep{roustant2020group}. Remark that this dataset is different from the one in \cite{zhang2020latent}. In this problem, there are various shapes with several filling configurations that can be considered as 12 different levels (see Figure \ref{fig:beam_bending_shapes}). 3 groups, each one corresponding to a given filling configuration. The underlying physical problem is the following: fix the beam at one end and apply force at the other. Under linear elasticity assumptions, the beam length at the cross section can be approximated by the following quantity \citep{gere1997sp}:
\begin{equation*}
f_{BB}(L,S,I) = \frac{L^3}{3S^2I}
\end{equation*}
where $L\in [10,20]$ and $S\in[1,2]$, see \cite{roustant2020group} for a more precise description of the variables. $I$ is treated as a categorical variable with the associated continuous values $(0.0833, 0.139, 0.380, 0.0796, 0.133, 0.363,
0.0859,\\ 0.136, 0.360, 0.0922, 0.138, 0.369)$ for all 12 shapes in Figure \ref{fig:beam_bending_shapes}. This gives a total of 2 continuous and 1 categorical variables. We consider 5 different sizes for the training sets 36/72/108/144/180 corresponding to 3/6/9/12/15 samples by level, respectively.

\begin{figure}[h!]
\begin{center}
\includegraphics[width=0.6\linewidth]{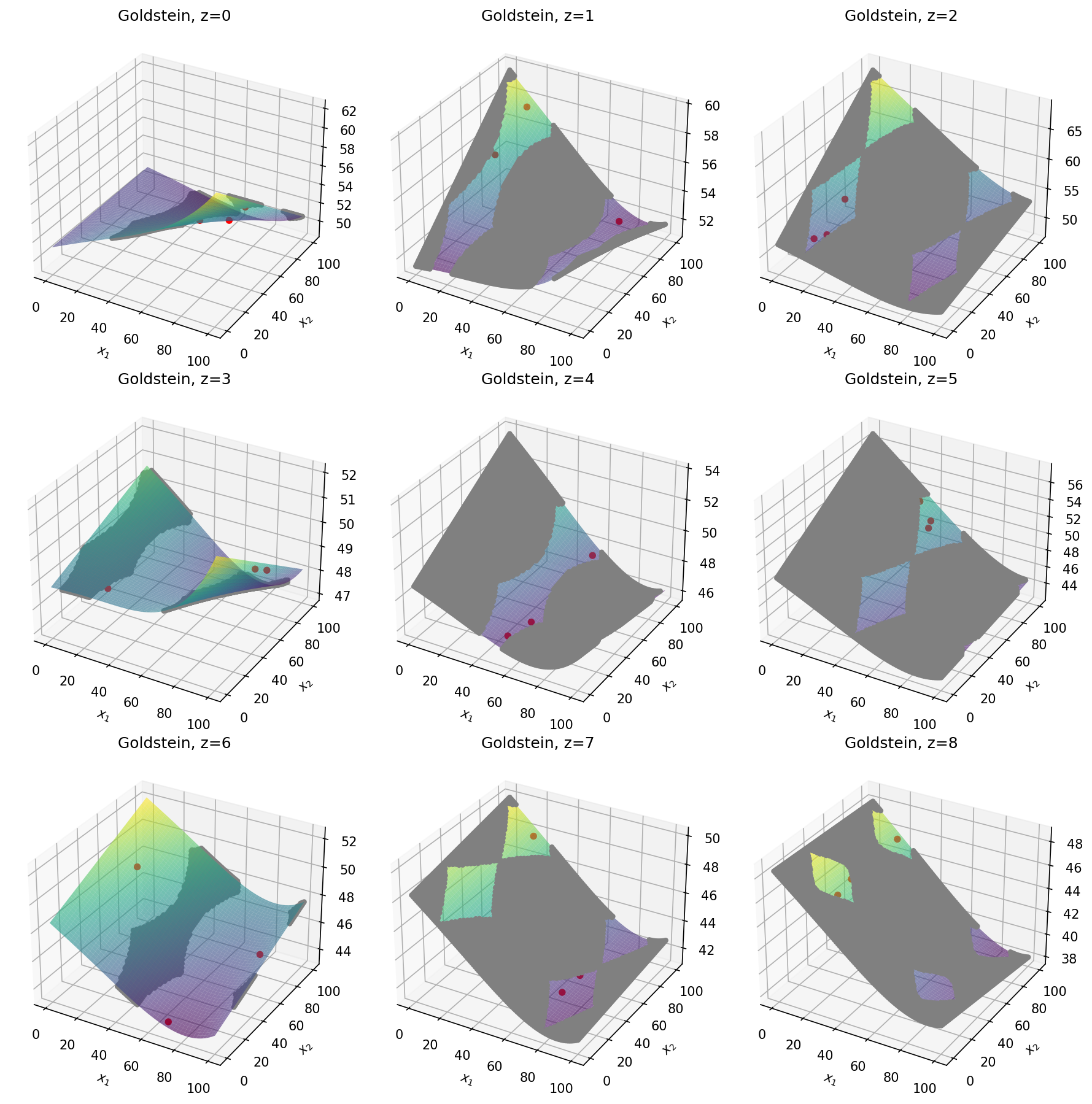}
\end{center}
\caption{Goldstein. Bullet points represent training samples. Contour plots give the value of the function for each level. \label{fig:goldstein}}
\end{figure}

\paragraph{Goldstein}~\\
The Goldstein function \citep{pelamatti2020mixed,picheny2013benchmark} is defined for two continuous variables and a categorical variable with 9 levels as follows:
\[
\begin{aligned}
f(x_1, x_2, z) = f(x_1, x_2, x_3, x_4) =\ & 53.3108 + 0.184901x_1 - 5.02914x_1^3 \cdot 10^{-6} + 7.72522x_1^4 \cdot 10^{-8} \\
& - 0.0870775x_2 - 0.106959x_3 + 7.98772x_3^3 \cdot 10^{-6} \\
& + 0.00242482x_4 + 1.32851x_4^3 \cdot 10^{-6} - 0.00146393x_1 x_2 \\
& - 0.00301588x_1 x_3 - 0.00272291x_1 x_4 + 0.0017004x_2 x_3 \\
& + 0.0038428x_2 x_4 - 0.000198969x_3 x_4 + 1.86025x_1 x_2 x_3 \cdot 10^{-5} \\
& - 1.88719x_1 x_2 x_4 \cdot 10^{-6} + 2.50923x_1 x_3 x_4 \cdot 10^{-5} \\
& - 5.62199x_2 x_3 x_4 \cdot 10^{-5}
\end{aligned}
\]
with $x_3=20, x_4=20$ if $z=0$, 
      $x_3=20, x_4=50$ if $z=1$,
      $x_3=20, x_4=80$ if $z=2$,
      $x_3=50, x_4=20$ if $z=3$, 
      $x_3=50, x_4=50$ if $z=4$,
      $x_3=50, x_4=80$ if $z=5$,
      $x_3=80, x_4=20$ if $z=6$, 
      $x_3=80, x_4=50$ if $z=7$,
      $x_3=80, x_4=80$ if $z=8$.
where $x_1, x_2\in [0,1]$, $z\in \llbracket 9 \rrbracket$. The constraint $g(x_1, x_2, z) = c_1 \sin(\frac{x_1}{10})^3 + c_2 \cos(\frac{x_2}{10})^2\leq 0$ is used to create the design of experiments,
with $c_1=2, c_2=0.5$ if $z=0$, 
      $c_1=2, c_2=-1$ if $z=1$,
      $c_1=2, c_2=-2$ if $z=2$,
      $c_1=-2, c_2=0.5$ if $z=3$, 
      $c_1=-2, c_2=-1$ if $z=4$,
      $c_1=-2, c_2=-2$ if $z=5$,
      $c_1=1, c_2=0.5$ if $z=6$, 
      $c_1=1, c_2=-1$ if $z=7$,
      $c_1=1, c_2=-2$ if $z=8$.
We consider 5 different sizes for the training sets /54/81/108/135 corresponding to 3/6/9/12/15 samples by level, respectively. Figure \ref{fig:goldstein} gives an illustration of the Goldstein function.

\section{Experimental setup}
\label{appendix:exp_setup}

In this section, details of the experiments are presented to ensure reproducibility. As a matter of fact, we found that several papers commented only on the parameters specific to their methods, and omitted critical details about the optimization.

\paragraph{Continuous kernel}~\\
We use an ARD RBF kernel for the continuous inputs. The RBF kernel is parameterized by the lengthscales. For each coordinate, the upper bound on the lengthscales is selected as two times the maximum distance between two input points in the training set and the lower bound is half the minimum distance between input points in the training set (coordinate-wise). We observed that fixing the minimal lengthscale to an arbitrary small value leads to failures in the optimization. In our code, we also propose another parameterization with the log of precisions giving similar results when setting the limits on the parameters correctly.

\paragraph{Categorical kernels}~\\
For the categorical part, the bounds on the parameters depend on the method. For the hypersphere variants, it is often difficult to know which are the bounds. In fact, we can consider hyperspheres allowing or not negative correlations. For the Hypersphere variants, we thus take an upper bound equal to $\frac{\pi}{2}$ for all methods except Ho\_NC and He\_NC where we allow negative correlations by taking an upper bound of $\pi$. When the kernel is a nested variant, we also use an upper bound of $\pi$ when between or within kernels are hyperspheres. For LVGP, latent representations have values in [-3,3]. Other categorical kernels do not require to fix their bounds.

\paragraph{Optimization}~\\
Parameters are selected via a maximization of the marginal log likelihood, where we use the closed-form formulas for the mean and global variance parameters of the GP (CF \cite{rasmussen2003gaussian}). Both the nugget and the parameters of the continuous and categorical kernels are optimized with the L-BFGS-B algorithm. The bounds on the nugget are $[1e-8,1e-4]$. We use 96 restart points obtained via a maximin LHS.

\paragraph{Scaling of the data}~\\
In practice, we considered versions of the datasets where both the continuous inputs and the outputs have been scaled to have a mean equal to zero and a variance equal to one.

\section{Performance profiles}
\label{appendix:performance_profiles}

\subsection{Presentation of the performance profiles}
Let us denote by $I$ the number of methods, $J$ the number of datasets, and $K$ the number of experiments. Let us consider the methods $i=1, \cdots, I$, and the datasets $j=1, \cdots, J$. We have access to several experiments $k=1, \cdots, K$. One specific experiment is associated to a specific design of experiments (and corresponds to one random seed). We observe the score $s_{i,j,k} \in \mathbb{R}$ for the method $i$ on the $k$-th experiment on the dataset $j$. Without loss of generality, we assume that the score must be minimized (for instance if the score is a RRMSE). Instead of looking at aggregated scores for each pair (dataset, method) according to all experiments, typically an average or median, we propose to consider each experiment separately.

In our experiments, all the (dataset, method) pairs share the same number of experiments. However, when some method is not applied to a given dataset, we arbitrarily set the score (for all experiments) to $\infty$. Given a method $i$, we can then denote by $\mathcal{J}_{i} = \{(j,k)\in \llbracket J \rrbracket \times \llbracket K \rrbracket : s_{i,j,k}<\infty\}$ the set of (dataset, experiment) pairs where the method $i$ is evaluated. Similarly, for a dataset $j$, we can denote by $\mathcal{I}_{j} = \{(i,k)\in \llbracket I \rrbracket \times \llbracket K \rrbracket : s_{i,j,k}<\infty\}$ the set of (method, experiment) pairs where the dataset $j$ is evaluated.

For a given dataset $j=1, \cdots, J$, we first sort all the scores obtained with all possible experiments and methods $\{s_{i,j,k} : (i,k)\in \mathcal{I}_{j} \}$. Given a quantile level $\tau \in [0,1]$, we then define the quantile of order $\tau$ of the latter set as $q_{j,\tau} := \mathrm{Quantile}_{\tau}( \{s_{i,j,k} : (i,k)\in \mathcal{I}_{j} \}) := \inf \left\{ t\in \mathbb{R} : \frac{1}{|\mathcal{I}_j|}\sum\limits_{(i,k)\in \mathcal{I}_j} \mathbb{1}_{\{s_{i,j,k}\leq t\}} \geq \tau \right\}$. This quantile in fact corresponds to the $\lceil \tau|\mathcal{I}_j|\rceil$-th smallest score value for the given dataset over all possible (experiment,method) pairs. 

For a given method $i=1,\cdots, I$, we can now measure its global performance thanks to the rankings of the experiments realized with this method on all the datasets. Choosing a performance level $\tau \in [0,1]$, we introduce the following quantity: $p_{i}(\tau) = \frac{1}{|\mathcal{J}_i|} |\{ (j,k)\in \mathcal{J}_i : s_{i,j,k} \leq q_{j,\tau}\}|$. The latter quantity evaluates the proportion of (experiment, dataset) pairs where the (experiment, method) pair is in the top $\tau$ performing methods of all experiments performed on the given dataset. The performance profile plot of a method is then characterized by the function $p_{i} : [0,1] \rightarrow [0,1]$.

\subsection{Additional performance profiles}

\begin{figure}[h!]
\begin{center}
\includegraphics[width=0.9\linewidth]{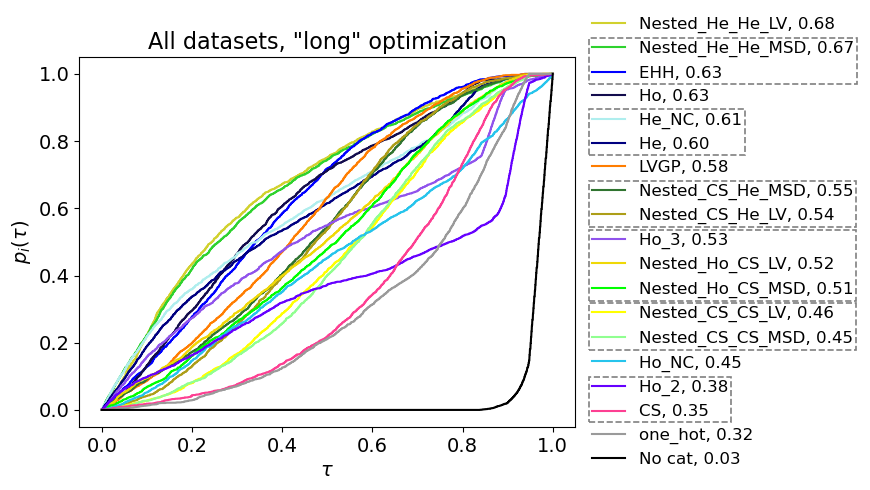}
\end{center}
\caption{Performance profiles using the RRMSE. All datasets and all kernels are considered. The score after the name of the method $i$ is $\textbf{AUC}(p_i)$, methods are sorted in AUC decreasing order.\label{fig:profile_all_RRMSE_FULL}}
\end{figure}

In Figure \ref{fig:profile_all_RRMSE_FULL}, we consider all datasets. In comparison to Figure \ref{fig:profile_all_RRMSE_noGK_LONG} using the same optimization options, the LVGP and MSD automatic variants of nested kernels with He/He achieve even better results and outperform hypersphere models. This additional result is not surprising, as datasets with a group structure are added to this performance profile.

\section{Clustering}
\label{appendix:clustering}
In this section, we give more details about the clustering procedure and provide additional results about the impact of group selection on the regression scores.

\subsection{Choice of the number of clusters}

A fundamental question when performing clustering is the selection of the number of clusters. \cite{roustant2020group} suggest setting the number of clusters by an exhaustive search. For all sizes $2\leq Q \leq C-1$, one can rely on cross-validation to compute a prediction criterion in order to define the best value of $Q$. This strategy can however be costly because it requires the training of several Gaussian processes. Instead, we propose to launch the categorical GP only once with a pre-determined number of clusters. The selection of the 'optimal' number of clusters remains a topic of ongoing debate, with no universally accepted criterion. Specific methods like Gaussian mixture models employ information criteria to penalize the complexity of the model, while methods such as the elbow plots are more general but harder to apply. Other methods such as the Density-Based Clustering Validation (DBCV) score \citep{moulavi2014density} are also not suitable, as they require a large number of points to make sense. We recommend using the Silhouette score \citep{rousseeuw1987silhouettes} which only requires pseudo-distances, unlike some criteria that manipulate data in Euclidean spaces. In the following, we describe the Silhouette score.

Let $2\leq Q \leq C-1$ be the number of clusters, and let $\mathcal{C}_1, \cdots, \mathcal{C}_{Q}$ be a partition of $\llbracket C \rrbracket$ into $Q$ clusters obtained by any clustering algorithm. The cluster of $z$ is denoted as $q(z)\in \llbracket Q \rrbracket$. We suppose we have access to a pseudo-distance $d: \llbracket C \rrbracket \times \llbracket C \rrbracket \rightarrow [0, +\infty)$. The Silhouette coefficient of $z\in \llbracket C \rrbracket$ is defined as:
\begin{equation*}
    s(z) = \frac{b(z)-a(z)}{\max(a(z), b(z))}
\end{equation*}
where $a(z) = \frac{1}{|\mathcal{C}_{q(z)}|-1}\sum\limits_{z'\in \mathcal{C}_{q(z)}\setminus \{z\}} d(z,z')$ is the average distance of the point to its group (with $a(z)=0$ if $|\mathcal{C}_{q(z)}|=1$), and $b(z) = \min\limits_{i\neq q(z)}\frac{1}{|\mathcal{C}_i|} \sum_{z'\in \mathcal{C}_i} d(z,z')$ is the average distance of the point from its neighbouring group. The global Silhouette score of the clustering is defined as follows:
\begin{equation*}
    s_{\mathrm{Silhouette}}( \{\mathcal{C}_1, \cdots, \mathcal{C_Q}\}) = \frac{1}{Q} \sum_{i=1}^Q \frac{1}{|\mathcal{C}_i|}\sum_{z\in \mathcal{C}_i} s(z)
\end{equation*}

The Silhouette score varies between -1 (worst classification) and 1 (best classification). As we can see from the definition of the Silhouette score, it is naturally only suitable for partitions having between $2$ and $C-1$ clusters. In the case where we want to consider $1$ or $C$ clusters, the model in fact corresponds to the CS or hypersphere models. In our experiments, we limit the choice between $2$ and $C-1$ for an automatic selection maximizing the Silhouette score. For a more general case, we encourage preliminary visualizations to identify the presence obvious groups, and, if not, run the Gaussian process regression with a hypersphere model.

\subsection{Regression scores depending on the group selection strategy.}

\begin{figure}[h!]
\begin{center}
\includegraphics[width=0.49\linewidth]{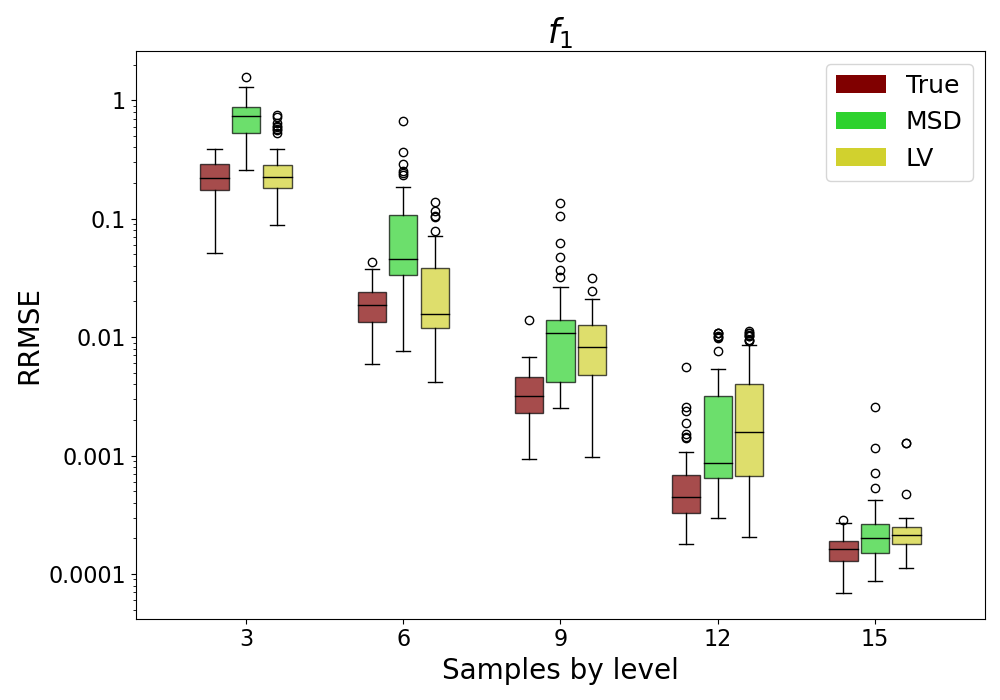}
\includegraphics[width=0.49\linewidth]{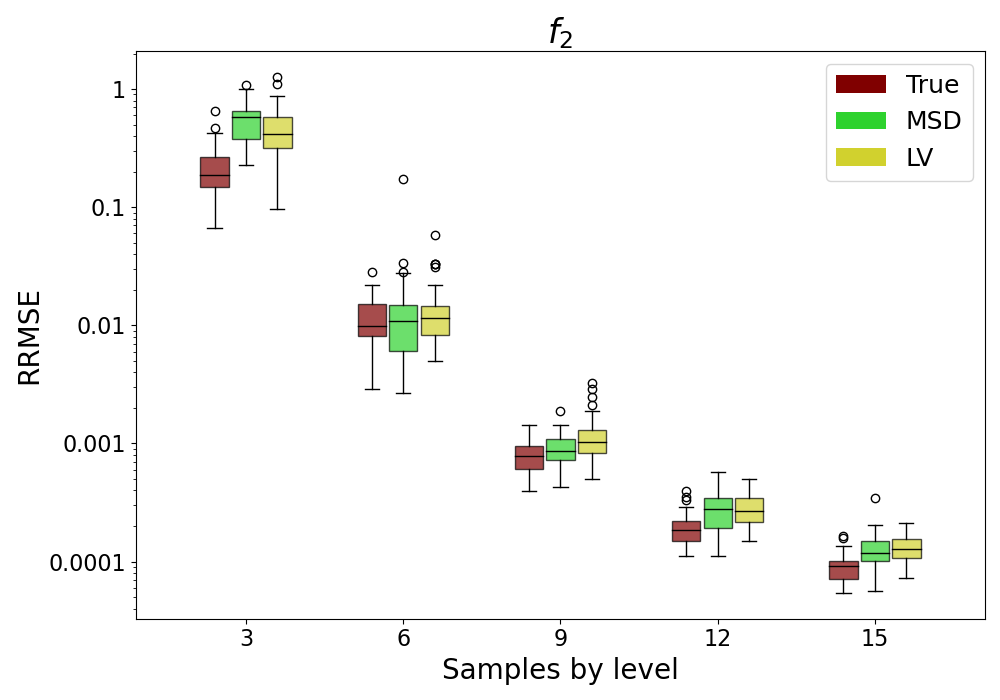}
\includegraphics[width=0.49\linewidth]{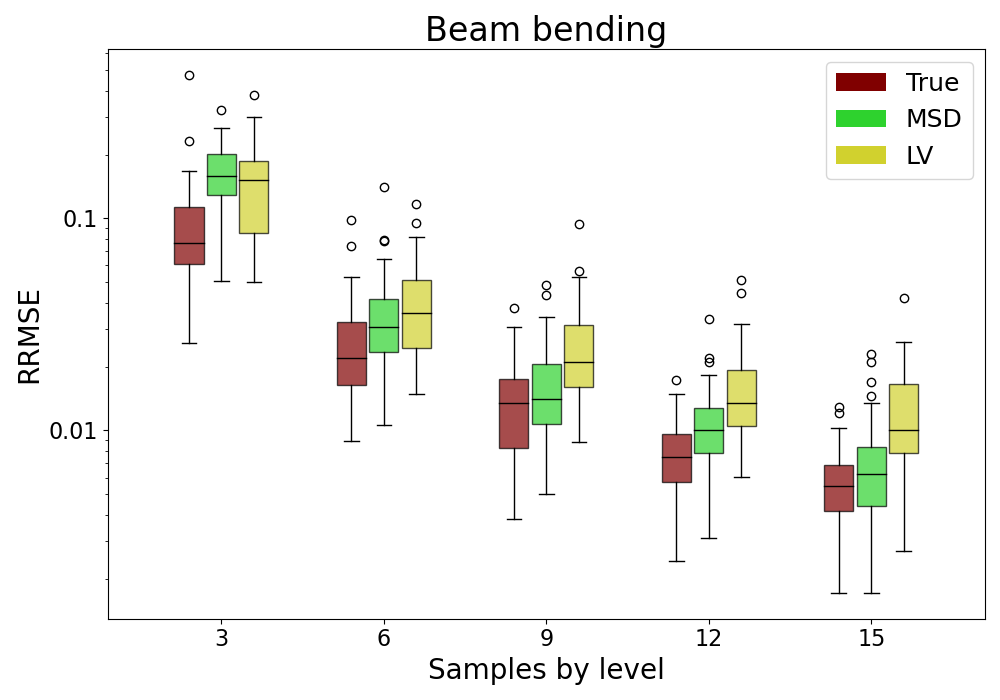}
\end{center}
\caption{Boxplots of the RRSME over the 50 experiments obtained with the Nested\_He\_He kernel relying on different group selection strategies. We consider different numbers of samples per level for the three datasets with known group structure.\label{fig:LV_vs_MSD_scores_three}}
\end{figure}

 In this section, we consider the RRMSE of the Gaussian process regression using the Nested\_He\_He graph kernels. We compare three group selection strategies on datasets with known group structures, including the true groups, and groups obtained thanks to clustering using respectively the MSD and LVGP representations. We give the boxplots of the RRMSE for the 50 experiments on the datasets $f_1$, $f_2$ and beam bending with various sizes in Figure \ref{fig:LV_vs_MSD_scores_three}. First of all, the overall error decreases with the number of samples per level in all cases, which is not surprising. In any case, the nested kernel version with real groups performs best. The automatic selection with LVGP or MSD obtain errors of the same order of magnitude as in the case where true groups are used for all datasets. Selection with MSD seems preferable or equivalent to that of LVGP when the number of samples per level is greater than 6 for $f_2$ and beam bending, and 9 for $f_1$. This supports the observation  of Section \ref{sec:inferring_group_structure} that a sufficient number of samples per level is necessary to take full advantage of representation with target encodings.

\section{Influence of the optimization}
\label{appendix:optimization}

\begin{figure}[h!]
    \centering
    \begin{subfigure}{0.49\textwidth}
        \centering
        \includegraphics[width=\linewidth]{boxplot_RRMSE_f1_GK_13_sbc6.png}
        \caption{"long" optimization}
        \label{fig:boxplots_groups_A}
    \end{subfigure}
    \hfill
    \begin{subfigure}{0.49\textwidth}
        \centering
        \includegraphics[width=\linewidth]{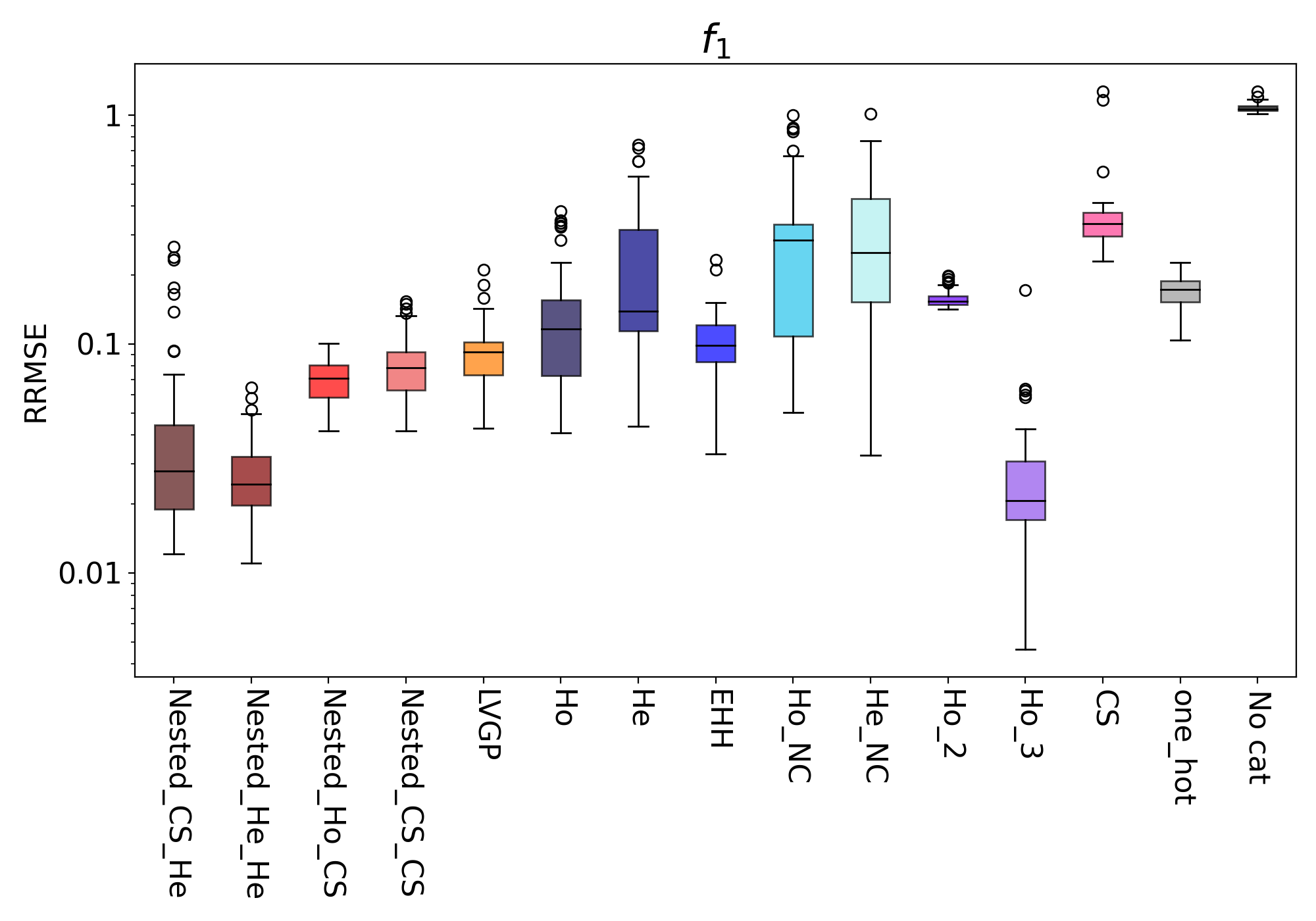}
        \caption{"short" optimization}
        \label{fig:boxplots_groups_B}
    \end{subfigure}
    \caption{Boxplots of the RRSME over the 50 experiments for all methods on the dataset $f_1$ with 6 samples by level depending on the optimization options.}
    \label{fig:boxplots_groups2}
\end{figure}

As we discussed in Section \ref{sec:experiments2}, optimization requires special attention and has a significant impact on the global performance of the models used. This impact is even more visible on hypersphere models, as can be seen in Figure \ref{fig:profile_all_RRMSE_noGK}. We add the comparison of boxplots the dataset $f_1$ with 6 samples by level in Figure \ref{fig:boxplots_groups2} for the "long" and "short" optimization. RRMSE are severely degraded for all the hypersphere models. Using the homoscedastic hypersphere kernel, performing the 96 restarts of the optimization takes a total time of 842 seconds for the "short" optimization while it takes 9620 seconds for the "long" optimization.

\end{appendix}

\clearpage

\bibliographystyle{elsarticle-harv}
\bibliography{bibliography}

\begin{thebibliography}{87}
\expandafter\ifx\csname natexlab\endcsname\relax\def\natexlab#1{#1}\fi
\providecommand{\url}[1]{\texttt{#1}}
\providecommand{\href}[2]{#2}
\providecommand{\path}[1]{#1}
\providecommand{\DOIprefix}{doi:}
\providecommand{\ArXivprefix}{arXiv:}
\providecommand{\URLprefix}{URL: }
\providecommand{\Pubmedprefix}{pmid:}
\providecommand{\doi}[1]{\href{http://dx.doi.org/#1}{\path{#1}}}
\providecommand{\Pubmed}[1]{\href{pmid:#1}{\path{#1}}}
\providecommand{\bibinfo}[2]{#2}
\ifx\xfnm\relax \def\xfnm[#1]{\unskip,\space#1}\fi
\bibitem[{Aitchison and Aitken(1976)}]{aitchison1976multivariate}
\bibinfo{author}{Aitchison, J.}, \bibinfo{author}{Aitken, C.G.},
  \bibinfo{year}{1976}.
\newblock \bibinfo{title}{Multivariate binary discrimination by the kernel
  method}.
\newblock \bibinfo{journal}{Biometrika} \bibinfo{volume}{63},
  \bibinfo{pages}{413--420}.
\bibitem[{Balandat et~al.(2020)Balandat, Karrer, Jiang, Daulton, Letham, Wilson
  and Bakshy}]{balandat2020botorch}
\bibinfo{author}{Balandat, M.}, \bibinfo{author}{Karrer, B.},
  \bibinfo{author}{Jiang, D.}, \bibinfo{author}{Daulton, S.},
  \bibinfo{author}{Letham, B.}, \bibinfo{author}{Wilson, A.G.},
  \bibinfo{author}{Bakshy, E.}, \bibinfo{year}{2020}.
\newblock \bibinfo{title}{Botorch: A framework for efficient monte-carlo
  bayesian optimization}.
\newblock \bibinfo{journal}{Advances in neural information processing systems}
  \bibinfo{volume}{33}, \bibinfo{pages}{21524--21538}.
\bibitem[{Bartoli et~al.(2019)Bartoli, Lefebvre, Dubreuil, Olivanti, Priem,
  Bons, Martins and Morlier}]{bartoli2019adaptive}
\bibinfo{author}{Bartoli, N.}, \bibinfo{author}{Lefebvre, T.},
  \bibinfo{author}{Dubreuil, S.}, \bibinfo{author}{Olivanti, R.},
  \bibinfo{author}{Priem, R.}, \bibinfo{author}{Bons, N.},
  \bibinfo{author}{Martins, J.R.}, \bibinfo{author}{Morlier, J.},
  \bibinfo{year}{2019}.
\newblock \bibinfo{title}{Adaptive modeling strategy for constrained global
  optimization with application to aerodynamic wing design}.
\newblock \bibinfo{journal}{Aerospace Science and technology}
  \bibinfo{volume}{90}, \bibinfo{pages}{85--102}.
\bibitem[{Baudin et~al.(2015)Baudin, Dutfoy, Iooss and
  Popelin}]{baudin2015open}
\bibinfo{author}{Baudin, M.}, \bibinfo{author}{Dutfoy, A.},
  \bibinfo{author}{Iooss, B.}, \bibinfo{author}{Popelin, A.L.},
  \bibinfo{year}{2015}.
\newblock \bibinfo{title}{Open turns: An industrial software for uncertainty
  quantification in simulation}.
\newblock \bibinfo{journal}{arXiv preprint arXiv:1501.05242} .
\bibitem[{Beauthier et~al.(2014)Beauthier, Mahajan, Sainvitu, Hendrick,
  Sharifzadeh and Verstraete}]{beauthier2014hypersonic}
\bibinfo{author}{Beauthier, C.}, \bibinfo{author}{Mahajan, A.},
  \bibinfo{author}{Sainvitu, C.}, \bibinfo{author}{Hendrick, P.},
  \bibinfo{author}{Sharifzadeh, S.}, \bibinfo{author}{Verstraete, D.},
  \bibinfo{year}{2014}.
\newblock \bibinfo{title}{Hypersonic cryogenic tank design using mixed-variable
  surrogate-based optimization}, in: \bibinfo{booktitle}{Engineering
  Optimization IV—Proceedings of the 4th International Conference on
  Engineering Optimization, ENGOPT}, pp. \bibinfo{pages}{543--549}.
\bibitem[{Ben-Ari and Steinberg(2007)}]{ben2007modeling}
\bibinfo{author}{Ben-Ari, E.N.}, \bibinfo{author}{Steinberg, D.M.},
  \bibinfo{year}{2007}.
\newblock \bibinfo{title}{Modeling data from computer experiments: an empirical
  comparison of kriging with mars and projection pursuit regression}.
\newblock \bibinfo{journal}{Quality Engineering} \bibinfo{volume}{19},
  \bibinfo{pages}{327--338}.
\bibitem[{Carpintero~Perez et~al.(2024)Carpintero~Perez, Da~Veiga, Garnier and
  Staber}]{perez2024gaussian}
\bibinfo{author}{Carpintero~Perez, R.}, \bibinfo{author}{Da~Veiga, S.},
  \bibinfo{author}{Garnier, J.}, \bibinfo{author}{Staber, B.},
  \bibinfo{year}{2024}.
\newblock \bibinfo{title}{{G}aussian process regression with sliced wasserstein
  weisfeiler-lehman graph kernels}, in: \bibinfo{booktitle}{International
  Conference on Artificial Intelligence and Statistics},
  \bibinfo{organization}{PMLR}. pp. \bibinfo{pages}{1297--1305}.
\bibitem[{Cuesta~Ramirez et~al.(2022)Cuesta~Ramirez, Le~Riche, Roustant,
  Perrin, Durantin and Gli{\`e}re}]{cuesta2022comparison}
\bibinfo{author}{Cuesta~Ramirez, J.}, \bibinfo{author}{Le~Riche, R.},
  \bibinfo{author}{Roustant, O.}, \bibinfo{author}{Perrin, G.},
  \bibinfo{author}{Durantin, C.}, \bibinfo{author}{Gli{\`e}re, A.},
  \bibinfo{year}{2022}.
\newblock \bibinfo{title}{A comparison of mixed-variables bayesian optimization
  approaches}.
\newblock \bibinfo{journal}{Advanced Modeling and Simulation in Engineering
  Sciences} \bibinfo{volume}{9}, \bibinfo{pages}{6}.
\bibitem[{Da~Veiga(2025)}]{da2025distributional}
\bibinfo{author}{Da~Veiga, S.}, \bibinfo{year}{2025}.
\newblock \bibinfo{title}{Distributional encoding for {G}aussian process
  regression with qualitative inputs}.
\newblock \bibinfo{journal}{arXiv preprint arXiv:2506.04813} .
\bibitem[{Da~Veiga et~al.(2021)Da~Veiga, Gamboa, Iooss and
  Prieur}]{da2021basics}
\bibinfo{author}{Da~Veiga, S.}, \bibinfo{author}{Gamboa, F.},
  \bibinfo{author}{Iooss, B.}, \bibinfo{author}{Prieur, C.},
  \bibinfo{year}{2021}.
\newblock \bibinfo{title}{Basics and trends in sensitivity analysis: Theory and
  practice in R}.
\newblock \bibinfo{publisher}{SIAM}.
\bibitem[{Deng et~al.(2017)Deng, Lin, Liu and Rowe}]{deng2017additive}
\bibinfo{author}{Deng, X.}, \bibinfo{author}{Lin, C.D.}, \bibinfo{author}{Liu,
  K.W.}, \bibinfo{author}{Rowe, R.K.}, \bibinfo{year}{2017}.
\newblock \bibinfo{title}{Additive {G}aussian process for computer models with
  qualitative and quantitative factors}.
\newblock \bibinfo{journal}{Technometrics} \bibinfo{volume}{59},
  \bibinfo{pages}{283--292}.
\bibitem[{Deshwal et~al.(2023)Deshwal, Ament, Balandat, Bakshy, Doppa and
  Eriksson}]{deshwal2023bayesian}
\bibinfo{author}{Deshwal, A.}, \bibinfo{author}{Ament, S.},
  \bibinfo{author}{Balandat, M.}, \bibinfo{author}{Bakshy, E.},
  \bibinfo{author}{Doppa, J.R.}, \bibinfo{author}{Eriksson, D.},
  \bibinfo{year}{2023}.
\newblock \bibinfo{title}{Bayesian optimization over high-dimensional
  combinatorial spaces via dictionary-based embeddings}, in:
  \bibinfo{booktitle}{International Conference on Artificial Intelligence and
  Statistics}, \bibinfo{organization}{PMLR}. pp. \bibinfo{pages}{7021--7039}.
\bibitem[{Deshwal et~al.(2021a)Deshwal, Belakaria and
  Doppa}]{deshwal2021bayesian}
\bibinfo{author}{Deshwal, A.}, \bibinfo{author}{Belakaria, S.},
  \bibinfo{author}{Doppa, J.R.}, \bibinfo{year}{2021}a.
\newblock \bibinfo{title}{Bayesian optimization over hybrid spaces}, in:
  \bibinfo{booktitle}{International Conference on Machine Learning},
  \bibinfo{organization}{PMLR}. pp. \bibinfo{pages}{2632--2643}.
\bibitem[{Deshwal et~al.(2021b)Deshwal, Belakaria and
  Doppa}]{deshwal2021mercer}
\bibinfo{author}{Deshwal, A.}, \bibinfo{author}{Belakaria, S.},
  \bibinfo{author}{Doppa, J.R.}, \bibinfo{year}{2021}b.
\newblock \bibinfo{title}{Mercer features for efficient combinatorial bayesian
  optimization}, in: \bibinfo{booktitle}{Proceedings of the AAAI Conference on
  Artificial Intelligence}, pp. \bibinfo{pages}{7210--7218}.
\bibitem[{Deshwal and Doppa(2021)}]{deshwal2021combining}
\bibinfo{author}{Deshwal, A.}, \bibinfo{author}{Doppa, J.},
  \bibinfo{year}{2021}.
\newblock \bibinfo{title}{Combining latent space and structured kernels for
  bayesian optimization over combinatorial spaces}.
\newblock \bibinfo{journal}{Advances in neural information processing systems}
  \bibinfo{volume}{34}, \bibinfo{pages}{8185--8200}.
\bibitem[{Deville et~al.(2024)Deville, Ginsbourger, Contributors, Durrande,
  Roustant, Rcpp, DiceKriging, Imports and Rcpp}]{deville2024package}
\bibinfo{author}{Deville, Y.}, \bibinfo{author}{Ginsbourger, D.},
  \bibinfo{author}{Contributors, O.R.}, \bibinfo{author}{Durrande, N.},
  \bibinfo{author}{Roustant, M.O.}, \bibinfo{author}{Rcpp, D.},
  \bibinfo{author}{DiceKriging, S.}, \bibinfo{author}{Imports, M.},
  \bibinfo{author}{Rcpp, L.}, \bibinfo{year}{2024}.
\newblock \bibinfo{title}{Package ‘kergp’}.
\newblock
  \bibinfo{howpublished}{\url{https://cran.r-project.org/web/packages/kergp/index.html}}.
\bibitem[{Dolan and Mor{\'e}(2002)}]{dolan2002benchmarking}
\bibinfo{author}{Dolan, E.D.}, \bibinfo{author}{Mor{\'e}, J.J.},
  \bibinfo{year}{2002}.
\newblock \bibinfo{title}{Benchmarking optimization software with performance
  profiles}.
\newblock \bibinfo{journal}{Mathematical programming} \bibinfo{volume}{91},
  \bibinfo{pages}{201--213}.
\bibitem[{Dreczkowski et~al.(2024)Dreczkowski, Grosnit and
  Bou~Ammar}]{dreczkowski2024framework}
\bibinfo{author}{Dreczkowski, K.}, \bibinfo{author}{Grosnit, A.},
  \bibinfo{author}{Bou~Ammar, H.}, \bibinfo{year}{2024}.
\newblock \bibinfo{title}{Framework and benchmarks for combinatorial and
  mixed-variable bayesian optimization}.
\newblock \bibinfo{journal}{Advances in Neural Information Processing Systems}
  \bibinfo{volume}{36}.
\bibitem[{Gardner et~al.(2018)Gardner, Pleiss, Weinberger, Bindel and
  Wilson}]{gardner2018gpytorch}
\bibinfo{author}{Gardner, J.}, \bibinfo{author}{Pleiss, G.},
  \bibinfo{author}{Weinberger, K.Q.}, \bibinfo{author}{Bindel, D.},
  \bibinfo{author}{Wilson, A.G.}, \bibinfo{year}{2018}.
\newblock \bibinfo{title}{Gpytorch: Blackbox matrix-matrix {G}aussian process
  inference with gpu acceleration}.
\newblock \bibinfo{journal}{Advances in neural information processing systems}
  \bibinfo{volume}{31}.
\bibitem[{Garrido-Merch{\'a}n and
  Hern{\'a}ndez-Lobato(2020)}]{garrido2020dealing}
\bibinfo{author}{Garrido-Merch{\'a}n, E.C.},
  \bibinfo{author}{Hern{\'a}ndez-Lobato, D.}, \bibinfo{year}{2020}.
\newblock \bibinfo{title}{Dealing with categorical and integer-valued variables
  in bayesian optimization with {G}aussian processes}.
\newblock \bibinfo{journal}{Neurocomputing} \bibinfo{volume}{380},
  \bibinfo{pages}{20--35}.
\bibitem[{Girard(2004)}]{girard2004approximate}
\bibinfo{author}{Girard, A.}, \bibinfo{year}{2004}.
\newblock \bibinfo{title}{Approximate methods for propagation of uncertainty
  with {G}aussian process models}.
\newblock \bibinfo{publisher}{University of Glasgow (United Kingdom)}.
\bibitem[{G{\'o}mez-Bombarelli et~al.(2018)G{\'o}mez-Bombarelli, Wei, Duvenaud,
  Hern{\'a}ndez-Lobato, S{\'a}nchez-Lengeling, Sheberla, Aguilera-Iparraguirre,
  Hirzel, Adams and Aspuru-Guzik}]{gomez2018automatic}
\bibinfo{author}{G{\'o}mez-Bombarelli, R.}, \bibinfo{author}{Wei, J.N.},
  \bibinfo{author}{Duvenaud, D.}, \bibinfo{author}{Hern{\'a}ndez-Lobato, J.M.},
  \bibinfo{author}{S{\'a}nchez-Lengeling, B.}, \bibinfo{author}{Sheberla, D.},
  \bibinfo{author}{Aguilera-Iparraguirre, J.}, \bibinfo{author}{Hirzel, T.D.},
  \bibinfo{author}{Adams, R.P.}, \bibinfo{author}{Aspuru-Guzik, A.},
  \bibinfo{year}{2018}.
\newblock \bibinfo{title}{Automatic chemical design using a data-driven
  continuous representation of molecules}.
\newblock \bibinfo{journal}{ACS central science} \bibinfo{volume}{4},
  \bibinfo{pages}{268--276}.
\bibitem[{Gopakumar et~al.(2018)Gopakumar, Gupta, Rana, Nguyen and
  Venkatesh}]{gopakumar2018algorithmic}
\bibinfo{author}{Gopakumar, S.}, \bibinfo{author}{Gupta, S.},
  \bibinfo{author}{Rana, S.}, \bibinfo{author}{Nguyen, V.},
  \bibinfo{author}{Venkatesh, S.}, \bibinfo{year}{2018}.
\newblock \bibinfo{title}{Algorithmic assurance: An active approach to
  algorithmic testing using bayesian optimisation}.
\newblock \bibinfo{journal}{Advances in Neural Information Processing Systems}
  \bibinfo{volume}{31}.
\bibitem[{Gower(1971)}]{gower1971general}
\bibinfo{author}{Gower, J.C.}, \bibinfo{year}{1971}.
\newblock \bibinfo{title}{A general coefficient of similarity and some of its
  properties}.
\newblock \bibinfo{journal}{Biometrics} \bibinfo{volume}{27},
  \bibinfo{pages}{857--871}.
\bibitem[{{GPy}(since 2012)}]{gpy2014}
\bibinfo{author}{{GPy}}, \bibinfo{year}{since 2012}.
\newblock \bibinfo{title}{{GPy}: A {G}aussian process framework in python}.
\newblock \bibinfo{howpublished}{\url{http://github.com/SheffieldML/GPy}}.
\bibitem[{Gramacy(2020)}]{gramacy2020surrogates}
\bibinfo{author}{Gramacy, R.B.}, \bibinfo{year}{2020}.
\newblock \bibinfo{title}{Surrogates: {G}aussian process modeling, design, and
  optimization for the applied sciences}.
\newblock \bibinfo{publisher}{Chapman and Hall/CRC}.
\bibitem[{Gretton et~al.(2012)Gretton, Borgwardt, Rasch, Sch{\"o}lkopf and
  Smola}]{mmd1}
\bibinfo{author}{Gretton, A.}, \bibinfo{author}{Borgwardt, K.M.},
  \bibinfo{author}{Rasch, M.J.}, \bibinfo{author}{Sch{\"o}lkopf, B.},
  \bibinfo{author}{Smola, A.}, \bibinfo{year}{2012}.
\newblock \bibinfo{title}{A kernel two-sample test}.
\newblock \bibinfo{journal}{The Journal of Machine Learning Research}
  \bibinfo{volume}{13}, \bibinfo{pages}{723--773}.
\bibitem[{Grosnit et~al.(2022)Grosnit, Malherbe, Tutunov, Wan, Wang and
  Ammar}]{grosnit2022boils}
\bibinfo{author}{Grosnit, A.}, \bibinfo{author}{Malherbe, C.},
  \bibinfo{author}{Tutunov, R.}, \bibinfo{author}{Wan, X.},
  \bibinfo{author}{Wang, J.}, \bibinfo{author}{Ammar, H.B.},
  \bibinfo{year}{2022}.
\newblock \bibinfo{title}{Boils: Bayesian optimisation for logic synthesis},
  in: \bibinfo{booktitle}{2022 Design, Automation \& Test in Europe Conference
  \& Exhibition (DATE)}, \bibinfo{organization}{IEEE}. pp.
  \bibinfo{pages}{1193--1196}.
\bibitem[{Halstrup(2016)}]{halstrup2016black}
\bibinfo{author}{Halstrup, M.}, \bibinfo{year}{2016}.
\newblock \bibinfo{title}{Black-box optimization of mixed discrete-continuous
  optimization problems}.
\newblock \bibinfo{journal}{Ph.D. thesis, TU Dortmund} .
\bibitem[{Jesus et~al.(2021)Jesus, Sohst, Vale and
  Suleman}]{jesus2021surrogate}
\bibinfo{author}{Jesus, T.}, \bibinfo{author}{Sohst, M.},
  \bibinfo{author}{Vale, J.L.d.}, \bibinfo{author}{Suleman, A.},
  \bibinfo{year}{2021}.
\newblock \bibinfo{title}{Surrogate based mdo of a canard configuration
  aircraft}.
\newblock \bibinfo{journal}{Structural and Multidisciplinary Optimization}
  \bibinfo{volume}{64}, \bibinfo{pages}{3747--3771}.
\bibitem[{Jones et~al.(1998)Jones, Schonlau and Welch}]{jones1998efficient}
\bibinfo{author}{Jones, D.R.}, \bibinfo{author}{Schonlau, M.},
  \bibinfo{author}{Welch, W.J.}, \bibinfo{year}{1998}.
\newblock \bibinfo{title}{Efficient global optimization of expensive black-box
  functions}.
\newblock \bibinfo{journal}{Journal of Global optimization}
  \bibinfo{volume}{13}, \bibinfo{pages}{455--492}.
\bibitem[{Kabalan et~al.(2025)Kabalan, Casenave, Bordeu and
  Ehrlacher}]{kabalan2025mmgp}
\bibinfo{author}{Kabalan, A.}, \bibinfo{author}{Casenave, F.},
  \bibinfo{author}{Bordeu, F.}, \bibinfo{author}{Ehrlacher, V.},
  \bibinfo{year}{2025}.
\newblock \bibinfo{title}{O-mmgp: Optimal mesh morphing {G}aussian process
  regression for solving pdes with non-parametric geometric variations}.
\newblock \bibinfo{journal}{arXiv preprint arXiv:2502.11632} .
\bibitem[{Karlsson et~al.(2020)Karlsson, Bliek, Verwer and
  de~Weerdt}]{karlsson2020continuous}
\bibinfo{author}{Karlsson, R.}, \bibinfo{author}{Bliek, L.},
  \bibinfo{author}{Verwer, S.}, \bibinfo{author}{de~Weerdt, M.},
  \bibinfo{year}{2020}.
\newblock \bibinfo{title}{Continuous surrogate-based optimization algorithms
  are well-suited for expensive discrete problems}, in:
  \bibinfo{booktitle}{Benelux Conference on Artificial Intelligence},
  \bibinfo{organization}{Springer}. pp. \bibinfo{pages}{48--63}.
\bibitem[{Katz(2011)}]{katz2011multivariable}
\bibinfo{author}{Katz, M.H.}, \bibinfo{year}{2011}.
\newblock \bibinfo{title}{Multivariable analysis: a practical guide for
  clinicians and public health researchers}.
\newblock \bibinfo{publisher}{Cambridge university press}.
\bibitem[{Kennedy and O'Hagan(2001)}]{kennedy2001bayesian}
\bibinfo{author}{Kennedy, M.C.}, \bibinfo{author}{O'Hagan, A.},
  \bibinfo{year}{2001}.
\newblock \bibinfo{title}{Bayesian calibration of computer models}.
\newblock \bibinfo{journal}{Journal of the Royal Statistical Society: Series B
  (Statistical Methodology)} \bibinfo{volume}{63}, \bibinfo{pages}{425--464}.
\bibitem[{Khan et~al.(2023)Khan, Cowen-Rivers, Grosnit, Robert, Greiff,
  Smorodina, Rawat, Akbar, Dreczkowski, Tutunov et~al.}]{khan2023toward}
\bibinfo{author}{Khan, A.}, \bibinfo{author}{Cowen-Rivers, A.I.},
  \bibinfo{author}{Grosnit, A.}, \bibinfo{author}{Robert, P.A.},
  \bibinfo{author}{Greiff, V.}, \bibinfo{author}{Smorodina, E.},
  \bibinfo{author}{Rawat, P.}, \bibinfo{author}{Akbar, R.},
  \bibinfo{author}{Dreczkowski, K.}, \bibinfo{author}{Tutunov, R.}, et~al.,
  \bibinfo{year}{2023}.
\newblock \bibinfo{title}{Toward real-world automated antibody design with
  combinatorial bayesian optimization}.
\newblock \bibinfo{journal}{Cell Reports Methods} \bibinfo{volume}{3}.
\bibitem[{Khuri and Good(1989)}]{khuri1989parameterization}
\bibinfo{author}{Khuri, A.I.}, \bibinfo{author}{Good, I.},
  \bibinfo{year}{1989}.
\newblock \bibinfo{title}{The parameterization of orthogonal matrices: A review
  mainly for statisticians}.
\newblock \bibinfo{journal}{South African Statistical Journal}
  \bibinfo{volume}{23}, \bibinfo{pages}{231--250}.
\bibitem[{Kim et~al.(2022)Kim, Choi and Cho}]{kim2022combinatorial}
\bibinfo{author}{Kim, J.}, \bibinfo{author}{Choi, S.}, \bibinfo{author}{Cho,
  M.}, \bibinfo{year}{2022}.
\newblock \bibinfo{title}{Combinatorial bayesian optimization with random
  mapping functions to convex polytopes}, in: \bibinfo{booktitle}{Uncertainty
  in Artificial Intelligence}, \bibinfo{organization}{PMLR}. pp.
  \bibinfo{pages}{1001--1011}.
\bibitem[{Kirchhoff and Kuhnt(2020)}]{kirchhoff2020gaussian}
\bibinfo{author}{Kirchhoff, D.}, \bibinfo{author}{Kuhnt, S.},
  \bibinfo{year}{2020}.
\newblock \bibinfo{title}{{G}aussian process models with low-rank correlation
  matrices for both continuous and categorical inputs}.
\newblock \bibinfo{journal}{arXiv preprint arXiv:2010.02574} .
\bibitem[{Kochanski et~al.(2017)Kochanski, Golovin, Karro, Solnik, Moitra and
  Sculley}]{kochanski2017bayesian}
\bibinfo{author}{Kochanski, G.}, \bibinfo{author}{Golovin, D.},
  \bibinfo{author}{Karro, J.}, \bibinfo{author}{Solnik, B.},
  \bibinfo{author}{Moitra, S.}, \bibinfo{author}{Sculley, D.},
  \bibinfo{year}{2017}.
\newblock \bibinfo{title}{Bayesian optimization for a better dessert}, in:
  \bibinfo{booktitle}{NIPS, workshop on Bayesian optimization}.
\bibitem[{Kondor and Lafferty(2002)}]{kondor2002diffusion}
\bibinfo{author}{Kondor, R.I.}, \bibinfo{author}{Lafferty, J.},
  \bibinfo{year}{2002}.
\newblock \bibinfo{title}{Diffusion kernels on graphs and other discrete
  structures}, in: \bibinfo{booktitle}{Proceedings of the 19th international
  conference on machine learning}, pp. \bibinfo{pages}{315--322}.
\bibitem[{Krige(1951)}]{krige1951statistical}
\bibinfo{author}{Krige, D.G.}, \bibinfo{year}{1951}.
\newblock \bibinfo{title}{A statistical approach to some basic mine valuation
  problems on the witwatersrand}.
\newblock \bibinfo{journal}{Journal of the Southern African Institute of Mining
  and Metallurgy} \bibinfo{volume}{52}, \bibinfo{pages}{119--139}.
\bibitem[{Lauvernet and Helbert(2020)}]{lauvernet2020metamodeling}
\bibinfo{author}{Lauvernet, C.}, \bibinfo{author}{Helbert, C.},
  \bibinfo{year}{2020}.
\newblock \bibinfo{title}{Metamodeling methods that incorporate qualitative
  variables for improved design of vegetative filter strips}.
\newblock \bibinfo{journal}{Reliability Engineering \& System Safety}
  \bibinfo{volume}{204}, \bibinfo{pages}{107083}.
\bibitem[{Liu and Nocedal(1989)}]{liu1989limited}
\bibinfo{author}{Liu, D.C.}, \bibinfo{author}{Nocedal, J.},
  \bibinfo{year}{1989}.
\newblock \bibinfo{title}{On the limited memory bfgs method for large scale
  optimization}.
\newblock \bibinfo{journal}{Mathematical programming} \bibinfo{volume}{45},
  \bibinfo{pages}{503--528}.
\bibitem[{Maus et~al.(2022)Maus, Jones, Moore, Kusner, Bradshaw and
  Gardner}]{maus2022local}
\bibinfo{author}{Maus, N.}, \bibinfo{author}{Jones, H.},
  \bibinfo{author}{Moore, J.}, \bibinfo{author}{Kusner, M.J.},
  \bibinfo{author}{Bradshaw, J.}, \bibinfo{author}{Gardner, J.},
  \bibinfo{year}{2022}.
\newblock \bibinfo{title}{Local latent space bayesian optimization over
  structured inputs}.
\newblock \bibinfo{journal}{Advances in neural information processing systems}
  \bibinfo{volume}{35}, \bibinfo{pages}{34505--34518}.
\bibitem[{McMillan et~al.(1999)McMillan, Sacks, Welch and
  Gao}]{mcmillan1999analysis}
\bibinfo{author}{McMillan, N.J.}, \bibinfo{author}{Sacks, J.},
  \bibinfo{author}{Welch, W.J.}, \bibinfo{author}{Gao, F.},
  \bibinfo{year}{1999}.
\newblock \bibinfo{title}{Analysis of protein activity data by {G}aussian
  stochastic process models}.
\newblock \bibinfo{journal}{Journal of Biopharmaceutical Statistics}
  \bibinfo{volume}{9}, \bibinfo{pages}{145--160}.
\bibitem[{Meunier et~al.(2022)Meunier, Pontil and
  Ciliberto}]{meunier2022distribution}
\bibinfo{author}{Meunier, D.}, \bibinfo{author}{Pontil, M.},
  \bibinfo{author}{Ciliberto, C.}, \bibinfo{year}{2022}.
\newblock \bibinfo{title}{Distribution regression with sliced wasserstein
  kernels}, in: \bibinfo{booktitle}{International Conference on Machine
  Learning}, \bibinfo{organization}{PMLR}. pp. \bibinfo{pages}{15501--15523}.
\bibitem[{Morris et~al.(1993)Morris, Mitchell and
  Ylvisaker}]{morris1993bayesian}
\bibinfo{author}{Morris, M.D.}, \bibinfo{author}{Mitchell, T.J.},
  \bibinfo{author}{Ylvisaker, D.}, \bibinfo{year}{1993}.
\newblock \bibinfo{title}{Bayesian design and analysis of computer experiments:
  use of derivatives in surface prediction}.
\newblock \bibinfo{journal}{Technometrics} \bibinfo{volume}{35},
  \bibinfo{pages}{243--255}.
\bibitem[{Moss et~al.(2020)Moss, Leslie, Beck, Gonzalez and
  Rayson}]{moss2020boss}
\bibinfo{author}{Moss, H.}, \bibinfo{author}{Leslie, D.},
  \bibinfo{author}{Beck, D.}, \bibinfo{author}{Gonzalez, J.},
  \bibinfo{author}{Rayson, P.}, \bibinfo{year}{2020}.
\newblock \bibinfo{title}{Boss: Bayesian optimization over string spaces}.
\newblock \bibinfo{journal}{Advances in neural information processing systems}
  \bibinfo{volume}{33}, \bibinfo{pages}{15476--15486}.
\bibitem[{Moulavi et~al.(2014)Moulavi, Jaskowiak, Campello, Zimek and
  Sander}]{moulavi2014density}
\bibinfo{author}{Moulavi, D.}, \bibinfo{author}{Jaskowiak, P.A.},
  \bibinfo{author}{Campello, R.J.}, \bibinfo{author}{Zimek, A.},
  \bibinfo{author}{Sander, J.}, \bibinfo{year}{2014}.
\newblock \bibinfo{title}{Density-based clustering validation}, in:
  \bibinfo{booktitle}{Proceedings of the 2014 SIAM international conference on
  data mining}, \bibinfo{organization}{SIAM}. pp. \bibinfo{pages}{839--847}.
\bibitem[{Murtagh and Contreras(2012)}]{murtagh2012algorithms}
\bibinfo{author}{Murtagh, F.}, \bibinfo{author}{Contreras, P.},
  \bibinfo{year}{2012}.
\newblock \bibinfo{title}{Algorithms for hierarchical clustering: an overview}.
\newblock \bibinfo{journal}{Wiley interdisciplinary reviews: data mining and
  knowledge discovery} \bibinfo{volume}{2}, \bibinfo{pages}{86--97}.
\bibitem[{Nguyen et~al.(2020)Nguyen, Gupta, Rana, Shilton and
  Venkatesh}]{nguyen2020bayesian}
\bibinfo{author}{Nguyen, D.}, \bibinfo{author}{Gupta, S.},
  \bibinfo{author}{Rana, S.}, \bibinfo{author}{Shilton, A.},
  \bibinfo{author}{Venkatesh, S.}, \bibinfo{year}{2020}.
\newblock \bibinfo{title}{Bayesian optimization for categorical and
  category-specific continuous inputs}, in: \bibinfo{booktitle}{Proceedings of
  the AAAI Conference on Artificial Intelligence}, pp.
  \bibinfo{pages}{5256--5263}.
\bibitem[{Notin et~al.(2021)Notin, Hern{\'a}ndez-Lobato and
  Gal}]{notin2021improving}
\bibinfo{author}{Notin, P.}, \bibinfo{author}{Hern{\'a}ndez-Lobato, J.M.},
  \bibinfo{author}{Gal, Y.}, \bibinfo{year}{2021}.
\newblock \bibinfo{title}{Improving black-box optimization in vae latent space
  using decoder uncertainty}.
\newblock \bibinfo{journal}{Advances in Neural Information Processing Systems}
  \bibinfo{volume}{34}, \bibinfo{pages}{802--814}.
\bibitem[{Oh et~al.(2021)Oh, Gavves and Welling}]{oh2021mixed}
\bibinfo{author}{Oh, C.}, \bibinfo{author}{Gavves, E.},
  \bibinfo{author}{Welling, M.}, \bibinfo{year}{2021}.
\newblock \bibinfo{title}{Mixed variable bayesian optimization with frequency
  modulated kernels}, in: \bibinfo{booktitle}{Uncertainty in Artificial
  Intelligence}, \bibinfo{organization}{PMLR}. pp. \bibinfo{pages}{950--960}.
\bibitem[{Oh et~al.(2019)Oh, Tomczak, Gavves and Welling}]{oh2019}
\bibinfo{author}{Oh, C.}, \bibinfo{author}{Tomczak, J.},
  \bibinfo{author}{Gavves, E.}, \bibinfo{author}{Welling, M.},
  \bibinfo{year}{2019}.
\newblock \bibinfo{title}{Combinatorial bayesian optimization using the graph
  cartesian product}.
\newblock \bibinfo{journal}{Advances in Neural Information Processing Systems}
  \bibinfo{volume}{32}.
\bibitem[{Oune and Bostanabad(2021)}]{oune2021latent}
\bibinfo{author}{Oune, N.}, \bibinfo{author}{Bostanabad, R.},
  \bibinfo{year}{2021}.
\newblock \bibinfo{title}{Latent map {G}aussian processes for mixed variable
  metamodeling}.
\newblock \bibinfo{journal}{Computer Methods in Applied Mechanics and
  Engineering} \bibinfo{volume}{387}, \bibinfo{pages}{114128}.
\bibitem[{Pelamatti(2020)}]{pelamatti2020mixed}
\bibinfo{author}{Pelamatti, J.}, \bibinfo{year}{2020}.
\newblock \bibinfo{title}{Mixed-variable Bayesian optimization: application to
  aerospace system design}.
\newblock Ph.D. thesis. Universit{\'e} de Lille.
\bibitem[{Pelamatti et~al.(2021)Pelamatti, Brevault, Balesdent, Talbi and
  Guerin}]{pelamatti2021mixed}
\bibinfo{author}{Pelamatti, J.}, \bibinfo{author}{Brevault, L.},
  \bibinfo{author}{Balesdent, M.}, \bibinfo{author}{Talbi, E.G.},
  \bibinfo{author}{Guerin, Y.}, \bibinfo{year}{2021}.
\newblock \bibinfo{title}{Mixed variable {G}aussian process-based surrogate
  modeling techniques: Application to aerospace design}.
\newblock \bibinfo{journal}{Journal of Aerospace Information Systems}
  \bibinfo{volume}{18}, \bibinfo{pages}{813--837}.
\bibitem[{Peyr{\'e} et~al.(2019)Peyr{\'e}, Cuturi et~al.}]{computational_ot}
\bibinfo{author}{Peyr{\'e}, G.}, \bibinfo{author}{Cuturi, M.}, et~al.,
  \bibinfo{year}{2019}.
\newblock \bibinfo{title}{Computational optimal transport: With applications to
  data science}.
\newblock \bibinfo{journal}{Foundations and Trends{\textregistered} in Machine
  Learning} \bibinfo{volume}{11}, \bibinfo{pages}{355--607}.
\bibitem[{Phillips and Venkatasubramanian(2011)}]{phillips2011gentle}
\bibinfo{author}{Phillips, J.M.}, \bibinfo{author}{Venkatasubramanian, S.},
  \bibinfo{year}{2011}.
\newblock \bibinfo{title}{A gentle introduction to the kernel distance}.
\newblock \bibinfo{journal}{arXiv preprint arXiv:1103.1625} .
\bibitem[{Picheny et~al.(2013)Picheny, Wagner and
  Ginsbourger}]{picheny2013benchmark}
\bibinfo{author}{Picheny, V.}, \bibinfo{author}{Wagner, T.},
  \bibinfo{author}{Ginsbourger, D.}, \bibinfo{year}{2013}.
\newblock \bibinfo{title}{A benchmark of kriging-based infill criteria for
  noisy optimization}.
\newblock \bibinfo{journal}{Structural and multidisciplinary optimization}
  \bibinfo{volume}{48}, \bibinfo{pages}{607--626}.
\bibitem[{Pinheiro and Bates(1996)}]{pinheiro1996unconstrained}
\bibinfo{author}{Pinheiro, J.C.}, \bibinfo{author}{Bates, D.M.},
  \bibinfo{year}{1996}.
\newblock \bibinfo{title}{Unconstrained parametrizations for
  variance-covariance matrices}.
\newblock \bibinfo{journal}{Statistics and computing} \bibinfo{volume}{6},
  \bibinfo{pages}{289--296}.
\bibitem[{Qian(2012)}]{qian2012sliced}
\bibinfo{author}{Qian, P.Z.}, \bibinfo{year}{2012}.
\newblock \bibinfo{title}{Sliced latin hypercube designs}.
\newblock \bibinfo{journal}{Journal of the American Statistical Association}
  \bibinfo{volume}{107}, \bibinfo{pages}{393--399}.
\bibitem[{Qian et~al.(2008)Qian, Wu and Wu}]{qian2008gaussian}
\bibinfo{author}{Qian, P.Z.G.}, \bibinfo{author}{Wu, H.}, \bibinfo{author}{Wu,
  C.J.}, \bibinfo{year}{2008}.
\newblock \bibinfo{title}{{G}aussian process models for computer experiments
  with qualitative and quantitative factors}.
\newblock \bibinfo{journal}{Technometrics} \bibinfo{volume}{50},
  \bibinfo{pages}{383--396}.
\bibitem[{Rapisarda et~al.(2007)Rapisarda, Brigo and
  Mercurio}]{rapisarda2007parameterizing}
\bibinfo{author}{Rapisarda, F.}, \bibinfo{author}{Brigo, D.},
  \bibinfo{author}{Mercurio, F.}, \bibinfo{year}{2007}.
\newblock \bibinfo{title}{Parameterizing correlations: a geometric
  interpretation}.
\newblock \bibinfo{journal}{IMA Journal of Management Mathematics}
  \bibinfo{volume}{18}, \bibinfo{pages}{55--73}.
\bibitem[{Rasmussen(2003)}]{rasmussen2003gaussian}
\bibinfo{author}{Rasmussen, C.E.}, \bibinfo{year}{2003}.
\newblock \bibinfo{title}{{G}aussian processes in machine learning}, in:
  \bibinfo{booktitle}{Summer school on machine learning}.
  \bibinfo{publisher}{Springer}, pp. \bibinfo{pages}{63--71}.
\bibitem[{Rousseeuw(1987)}]{rousseeuw1987silhouettes}
\bibinfo{author}{Rousseeuw, P.J.}, \bibinfo{year}{1987}.
\newblock \bibinfo{title}{Silhouettes: a graphical aid to the interpretation
  and validation of cluster analysis}.
\newblock \bibinfo{journal}{Journal of computational and applied mathematics}
  \bibinfo{volume}{20}, \bibinfo{pages}{53--65}.
\bibitem[{Roustant et~al.(2020)Roustant, Padonou, Deville, Clément, Perrin,
  Giorla and Wynn}]{roustant2020group}
\bibinfo{author}{Roustant, O.}, \bibinfo{author}{Padonou, E.},
  \bibinfo{author}{Deville, Y.}, \bibinfo{author}{Clément, A.},
  \bibinfo{author}{Perrin, G.}, \bibinfo{author}{Giorla, J.},
  \bibinfo{author}{Wynn, H.}, \bibinfo{year}{2020}.
\newblock \bibinfo{title}{Group kernels for {G}aussian process metamodels with
  categorical inputs}.
\newblock \bibinfo{journal}{SIAM/ASA Journal on Uncertainty Quantification}
  \bibinfo{volume}{8}, \bibinfo{pages}{775--806}.
\bibitem[{Ru et~al.(2020)Ru, Alvi, Nguyen, Osborne and
  Roberts}]{ru2020bayesian}
\bibinfo{author}{Ru, B.}, \bibinfo{author}{Alvi, A.}, \bibinfo{author}{Nguyen,
  V.}, \bibinfo{author}{Osborne, M.A.}, \bibinfo{author}{Roberts, S.},
  \bibinfo{year}{2020}.
\newblock \bibinfo{title}{Bayesian optimisation over multiple continuous and
  categorical inputs}, in: \bibinfo{booktitle}{International Conference on
  Machine Learning}, \bibinfo{organization}{PMLR}. pp.
  \bibinfo{pages}{8276--8285}.
\bibitem[{Sack et~al.(1989)Sack, Welch, Mitchell and Wynn}]{sack1989design}
\bibinfo{author}{Sack, J.}, \bibinfo{author}{Welch, W.},
  \bibinfo{author}{Mitchell, T.}, \bibinfo{author}{Wynn, H.},
  \bibinfo{year}{1989}.
\newblock \bibinfo{title}{Design and analysis of computer experiments (with
  discussion)}.
\newblock \bibinfo{journal}{Statistical Science} \bibinfo{volume}{4},
  \bibinfo{pages}{409--435}.
\bibitem[{Santner et~al.(2003)Santner, Williams, Notz and
  Williams}]{santner2003design}
\bibinfo{author}{Santner, T.J.}, \bibinfo{author}{Williams, B.J.},
  \bibinfo{author}{Notz, W.I.}, \bibinfo{author}{Williams, B.J.},
  \bibinfo{year}{2003}.
\newblock \bibinfo{title}{The design and analysis of computer experiments}.
  volume~\bibinfo{volume}{1}.
\newblock \bibinfo{publisher}{Springer}.
\bibitem[{Saves(2024)}]{saves2024high}
\bibinfo{author}{Saves, P.}, \bibinfo{year}{2024}.
\newblock \bibinfo{title}{High-dimensional multidisciplinary design
  optimization for aircraft eco-design}.
\newblock Ph.D. thesis. ISAE-SUPAERO.
\bibitem[{Saves et~al.(2022)Saves, Bartoli, Diouane, Lefebvre, Morlier, David,
  Nguyen~Van and Defoort}]{saves2022bayesian}
\bibinfo{author}{Saves, P.}, \bibinfo{author}{Bartoli, N.},
  \bibinfo{author}{Diouane, Y.}, \bibinfo{author}{Lefebvre, T.},
  \bibinfo{author}{Morlier, J.}, \bibinfo{author}{David, C.},
  \bibinfo{author}{Nguyen~Van, E.}, \bibinfo{author}{Defoort, S.},
  \bibinfo{year}{2022}.
\newblock \bibinfo{title}{Bayesian optimization for mixed variables using an
  adaptive dimension reduction process: applications to aircraft design}, in:
  \bibinfo{booktitle}{AIAA SciTech 2022 Forum}, p. \bibinfo{pages}{0082}.
\bibitem[{Saves et~al.(2023)Saves, Diouane, Bartoli, Lefebvre and
  Morlier}]{saves2023mixed}
\bibinfo{author}{Saves, P.}, \bibinfo{author}{Diouane, Y.},
  \bibinfo{author}{Bartoli, N.}, \bibinfo{author}{Lefebvre, T.},
  \bibinfo{author}{Morlier, J.}, \bibinfo{year}{2023}.
\newblock \bibinfo{title}{A mixed-categorical correlation kernel for {G}aussian
  process}.
\newblock \bibinfo{journal}{Neurocomputing} \bibinfo{volume}{550},
  \bibinfo{pages}{126472}.
\bibitem[{Saves et~al.(2024)Saves, Lafage, Bartoli, Diouane, Bussemaker,
  Lefebvre, Hwang, Morlier and Martins}]{saves2024smt}
\bibinfo{author}{Saves, P.}, \bibinfo{author}{Lafage, R.},
  \bibinfo{author}{Bartoli, N.}, \bibinfo{author}{Diouane, Y.},
  \bibinfo{author}{Bussemaker, J.}, \bibinfo{author}{Lefebvre, T.},
  \bibinfo{author}{Hwang, J.T.}, \bibinfo{author}{Morlier, J.},
  \bibinfo{author}{Martins, J.R.}, \bibinfo{year}{2024}.
\newblock \bibinfo{title}{Smt 2.0: A surrogate modeling toolbox with a focus on
  hierarchical and mixed variables {G}aussian processes}.
\newblock \bibinfo{journal}{Advances in Engineering Software}
  \bibinfo{volume}{188}, \bibinfo{pages}{103571}.
\bibitem[{Shepard et~al.(2015)Shepard, Brozell and
  Gidofalvi}]{shepard2015representation}
\bibinfo{author}{Shepard, R.}, \bibinfo{author}{Brozell, S.R.},
  \bibinfo{author}{Gidofalvi, G.}, \bibinfo{year}{2015}.
\newblock \bibinfo{title}{The representation and parametrization of orthogonal
  matrices}.
\newblock \bibinfo{journal}{The Journal of Physical Chemistry A}
  \bibinfo{volume}{119}, \bibinfo{pages}{7924--7939}.
\bibitem[{Snoek et~al.(2012)Snoek, Larochelle and Adams}]{snoek2012}
\bibinfo{author}{Snoek, J.}, \bibinfo{author}{Larochelle, H.},
  \bibinfo{author}{Adams, R.P.}, \bibinfo{year}{2012}.
\newblock \bibinfo{title}{Practical bayesian optimization of machine learning
  algorithms}.
\newblock \bibinfo{journal}{Advances in neural information processing systems}
  \bibinfo{volume}{25}.
\bibitem[{Stein(1987)}]{stein1987large}
\bibinfo{author}{Stein, M.}, \bibinfo{year}{1987}.
\newblock \bibinfo{title}{Large sample properties of simulations using latin
  hypercube sampling}.
\newblock \bibinfo{journal}{Technometrics} \bibinfo{volume}{29},
  \bibinfo{pages}{143--151}.
\bibitem[{Timoshenko and Gere(1997)}]{gere1997sp}
\bibinfo{author}{Timoshenko, S.}, \bibinfo{author}{Gere, J.},
  \bibinfo{year}{1997}.
\newblock \bibinfo{title}{Mechanics of materials}.
\newblock \bibinfo{journal}{PWS} \bibinfo{volume}{912}, \bibinfo{pages}{9}.
\bibitem[{Virtanen et~al.(2020)Virtanen, Gommers, Oliphant, Haberland, Reddy,
  Cournapeau, Burovski, Peterson, Weckesser, Bright, {van der Walt}, Brett,
  Wilson, Millman, Mayorov, Nelson, Jones, Kern, Larson, Carey, Polat, Feng,
  Moore, {VanderPlas}, Laxalde, Perktold, Cimrman, Henriksen, Quintero, Harris,
  Archibald, Ribeiro, Pedregosa, {van Mulbregt} and {SciPy 1.0
  Contributors}}]{2020SciPy-NMeth}
\bibinfo{author}{Virtanen, P.}, \bibinfo{author}{Gommers, R.},
  \bibinfo{author}{Oliphant, T.E.}, \bibinfo{author}{Haberland, M.},
  \bibinfo{author}{Reddy, T.}, \bibinfo{author}{Cournapeau, D.},
  \bibinfo{author}{Burovski, E.}, \bibinfo{author}{Peterson, P.},
  \bibinfo{author}{Weckesser, W.}, \bibinfo{author}{Bright, J.},
  \bibinfo{author}{{van der Walt}, S.J.}, \bibinfo{author}{Brett, M.},
  \bibinfo{author}{Wilson, J.}, \bibinfo{author}{Millman, K.J.},
  \bibinfo{author}{Mayorov, N.}, \bibinfo{author}{Nelson, A.R.J.},
  \bibinfo{author}{Jones, E.}, \bibinfo{author}{Kern, R.},
  \bibinfo{author}{Larson, E.}, \bibinfo{author}{Carey, C.J.},
  \bibinfo{author}{Polat, {\.I}.}, \bibinfo{author}{Feng, Y.},
  \bibinfo{author}{Moore, E.W.}, \bibinfo{author}{{VanderPlas}, J.},
  \bibinfo{author}{Laxalde, D.}, \bibinfo{author}{Perktold, J.},
  \bibinfo{author}{Cimrman, R.}, \bibinfo{author}{Henriksen, I.},
  \bibinfo{author}{Quintero, E.A.}, \bibinfo{author}{Harris, C.R.},
  \bibinfo{author}{Archibald, A.M.}, \bibinfo{author}{Ribeiro, A.H.},
  \bibinfo{author}{Pedregosa, F.}, \bibinfo{author}{{van Mulbregt}, P.},
  \bibinfo{author}{{SciPy 1.0 Contributors}}, \bibinfo{year}{2020}.
\newblock \bibinfo{title}{{{SciPy} 1.0: Fundamental Algorithms for Scientific
  Computing in Python}}.
\newblock \bibinfo{journal}{Nature Methods} \bibinfo{volume}{17},
  \bibinfo{pages}{261--272}.
\newblock \DOIprefix\doi{10.1038/s41592-019-0686-2}.
\bibitem[{Wan et~al.(2021)Wan, Nguyen, Ha, Ru, Lu and Osborne}]{wan2021think}
\bibinfo{author}{Wan, X.}, \bibinfo{author}{Nguyen, V.}, \bibinfo{author}{Ha,
  H.}, \bibinfo{author}{Ru, B.}, \bibinfo{author}{Lu, C.},
  \bibinfo{author}{Osborne, M.A.}, \bibinfo{year}{2021}.
\newblock \bibinfo{title}{Think global and act local: Bayesian optimisation
  over high-dimensional categorical and mixed search spaces}.
\newblock \bibinfo{journal}{arXiv preprint arXiv:2102.07188} .
\bibitem[{Watanabe(2023)}]{watanabe2023}
\bibinfo{author}{Watanabe, S.}, \bibinfo{year}{2023}.
\newblock \bibinfo{title}{Tree-structured parzen estimator: Understanding its
  algorithm components and their roles for better empirical performance}.
\newblock \bibinfo{journal}{arXiv preprint arXiv:2304.11127} .
\bibitem[{Wistuba et~al.(2019)Wistuba, Rawat and Pedapati}]{wistuba2019}
\bibinfo{author}{Wistuba, M.}, \bibinfo{author}{Rawat, A.},
  \bibinfo{author}{Pedapati, T.}, \bibinfo{year}{2019}.
\newblock \bibinfo{title}{A survey on neural architecture search}.
\newblock \bibinfo{journal}{arXiv preprint arXiv:1905.01392} .
\bibitem[{Yin and Du(2022)}]{yin2022uncertainty}
\bibinfo{author}{Yin, J.}, \bibinfo{author}{Du, X.}, \bibinfo{year}{2022}.
\newblock \bibinfo{title}{Uncertainty quantification by convolutional neural
  network {G}aussian process regression with image and numerical data}, in:
  \bibinfo{booktitle}{AIAA SCITECH 2022 Forum}, p. \bibinfo{pages}{1100}.
\bibitem[{Zhang and Notz(2015)}]{zhang2015computer}
\bibinfo{author}{Zhang, Y.}, \bibinfo{author}{Notz, W.I.},
  \bibinfo{year}{2015}.
\newblock \bibinfo{title}{Computer experiments with qualitative and
  quantitative variables: A review and reexamination}.
\newblock \bibinfo{journal}{Quality Engineering} \bibinfo{volume}{27},
  \bibinfo{pages}{2--13}.
\bibitem[{Zhang et~al.(2020)Zhang, Tao, Chen and Apley}]{zhang2020latent}
\bibinfo{author}{Zhang, Y.}, \bibinfo{author}{Tao, S.}, \bibinfo{author}{Chen,
  W.}, \bibinfo{author}{Apley, D.W.}, \bibinfo{year}{2020}.
\newblock \bibinfo{title}{A latent variable approach to {G}aussian process
  modeling with qualitative and quantitative factors}.
\newblock \bibinfo{journal}{Technometrics} \bibinfo{volume}{62},
  \bibinfo{pages}{291--302}.
\bibitem[{Zhou et~al.(2011)Zhou, Qian and Zhou}]{zhou2011simple}
\bibinfo{author}{Zhou, Q.}, \bibinfo{author}{Qian, P.Z.},
  \bibinfo{author}{Zhou, S.}, \bibinfo{year}{2011}.
\newblock \bibinfo{title}{A simple approach to emulation for computer models
  with qualitative and quantitative factors}.
\newblock \bibinfo{journal}{Technometrics} \bibinfo{volume}{53},
  \bibinfo{pages}{266--273}.

\end{thebibliography}

\end{document}